\documentclass[10pt,twocolumn,letterpaper]{article}

\usepackage[pagenumbers]{cvpr} % To force page numbers, e.g. for an arXiv version

\usepackage{graphicx}
\usepackage{amsmath}
\usepackage{amssymb}
\usepackage{booktabs}
\usepackage{arydshln}
\usepackage{comment}
\usepackage{xcolor}
\definecolor{citecolor}{HTML}{0071BC}
\definecolor{linkcolor}{HTML}{ED1C24}
\usepackage[pagebackref=false,breaklinks=true,letterpaper=true,colorlinks,urlcolor=linkcolor,citecolor=citecolor,linkcolor=linkcolor,bookmarks=false]{hyperref}

\usepackage[capitalize]{cleveref}
\crefname{section}{Sec.}{Secs.}
\Crefname{section}{Section}{Sections}
\Crefname{table}{Table}{Tables}
\crefname{table}{Tab.}{Tabs.}

\def\cvprPaperID{819} % *** Enter the CVPR Paper ID here
\def\confName{CVPR}
\def\confYear{2023}

\begin{document}

%%%%%%%%% TITLE - PLEASE UPDATE
\title{Hybrid Neural Rendering for Large-Scale Scenes with Motion Blur} 
% Using Hybrid Neural Radiance Fields and Blur Simulations}

\author{Peng Dai$^{1}$\thanks{Equal contribution}
\hspace{1cm} Yinda Zhang$^{2*}$ \hspace{1cm} Xin Yu$^{1}$ \hspace{1cm} Xiaoyang Lyu$^{1}$ \hspace{1cm} Xiaojuan Qi$^{1}$\thanks{Corresponding author}\\
 $^1$The University of Hong Kong \hspace{1cm} $^2$ Google
% For a paper whose authors are all at the same institution,
% omit the following lines up until the closing ``}''.
% Additional authors and addresses can be added with ``\and'',
% just like the second author.
% To save space, use either the email address or home page, not both
}
\maketitle

\begin{abstract}
Rendering novel view images is highly desirable for many applications. Despite recent progress, it remains challenging to render high-fidelity and view-consistent novel views of large-scale scenes from in-the-wild images with inevitable artifacts (e.g., motion blur).
%Different from object-level rendering, the large-scale scene contains more information and details, thus requires the neural rendering model to have great capability.
To this end, we develop a hybrid neural rendering model that makes image-based representation and neural 3D representation join forces to render high-quality, view-consistent images. Moreover, images captured in the wild inevitably contain artifacts, such as motion blur, which deteriorates the quality of rendered images. Accordingly, we propose strategies to simulate blur effects on the rendered images to mitigate the negative influence of blurriness images and reduce their importance during training based on precomputed quality-aware weights.
 %Eventually, our approach achieves good performance by collaboratively working on both model and data perspectives. 
Extensive experiments on real and synthetic data demonstrate our model surpasses state-of-the-art point-based methods for novel view synthesis. The code is available at
{\url{https://daipengwa.github.io/Hybrid-Rendering-ProjectPage/}.}
    
\end{abstract}
\definecolor{ToRephaseColor}{rgb}{1.0, 0.4, 0.0}
\definecolor{TodoColor}{rgb}{1.0, 0.0, 1.0}
\definecolor{GreenColor}{rgb}{0.0, 0.0, 1.0}
\definecolor{PurpleColor}{rgb}{0.5, 0.0, 0.5}
\definecolor{BlueColor}{rgb}{0.0, 0.0, 1.0}
\definecolor{BabyBlueColor}{rgb}{0.5, 0.5, 1}
\definecolor{RevisionFixedColor}{rgb}{0.6, 0.0, 0.4}
\definecolor{FinalColor}{rgb}{0.0, 0.0, 0.0}

\def\sectionautorefname{Section}%
\def\subsectionautorefname{section}%
\def\subsubsectionautorefname{section}%

\newcommand{\bmat}[1]{\begin{bmatrix}#1\end{bmatrix}}
\newcommand{\pmat}[1]{\begin{pmatrix}#1\end{pmatrix}}

\newcommand{\todo}[1]{{\color{TodoColor} [TODO: #1]}}
\newcommand{\torephrase}[1]{ {\color{ToRephaseColor} [ToRephrase:] #1} }
\newcommand{\yd}[1]{\textcolor[rgb]{1,0.6,0.1}{{[YZ: #1]}}}
\newcommand{\xjqi}[1]{\textcolor[rgb]{0.9,0.3,0.3}{{[xjqi: #1]}}}

\newcommand{\degree}{\ensuremath{^{\circ}} }
\newcommand{\ceil}[1]{{\lceil #1 \rceil}}
\newcommand{\floor}[1]{{\lfloor #1 \rfloor}}

\newcommand{\Log}[1]{\log\left(#1\right)}
\newcommand{\NormTwo}[1]{\left\lVert#1\right\rVert_{2}}
\newcommand{\Norm}[1]{\left\lVert#1\right\rVert}
\newcommand{\mathbfit}[1]{\textbf{\textit{#1}}}
\newcommand{\Origin}{\mathbfit{O}}

\section{Introduction}
\begin{comment}
\xjqi{some thoughts: image-based rendering can directly take features from captured images and thus can better leverage high-quality cues in images; neural radiance field encode a scene as a whole and thus offers better consistency}

\xjqi{some thoughts on introduction stories: \textbf{P1}. our problem neural rendering of large-scale scenes in the wild. highlight the challenges: real-world data capture (unsatisfactory data and blur), insufficiency of using a neural radiance field to represent a model as a whole. \textbf{P2+P3}: research progress: major streams of work categorized into two lines of methods: 1.  PointNerf and so on which incorporates 3D representations into the neural rendering formulation, which advantages and disadvantages. 2. image-based rendering advantages and disadvantages. the above works aim to enrich the representation power of rendering leaving the data issue unresolved. Mention deblur nerf [why it doesn't apply to scannet??] and its limitations.\textbf{P4:} Our target address representation and data blur issue. work collectively to make neural rendering perform well in the wild. \textbf {P5:} highlight our contribution }
\end{comment}

\begin{figure}
    \centering
     \includegraphics[width=1.0\linewidth]{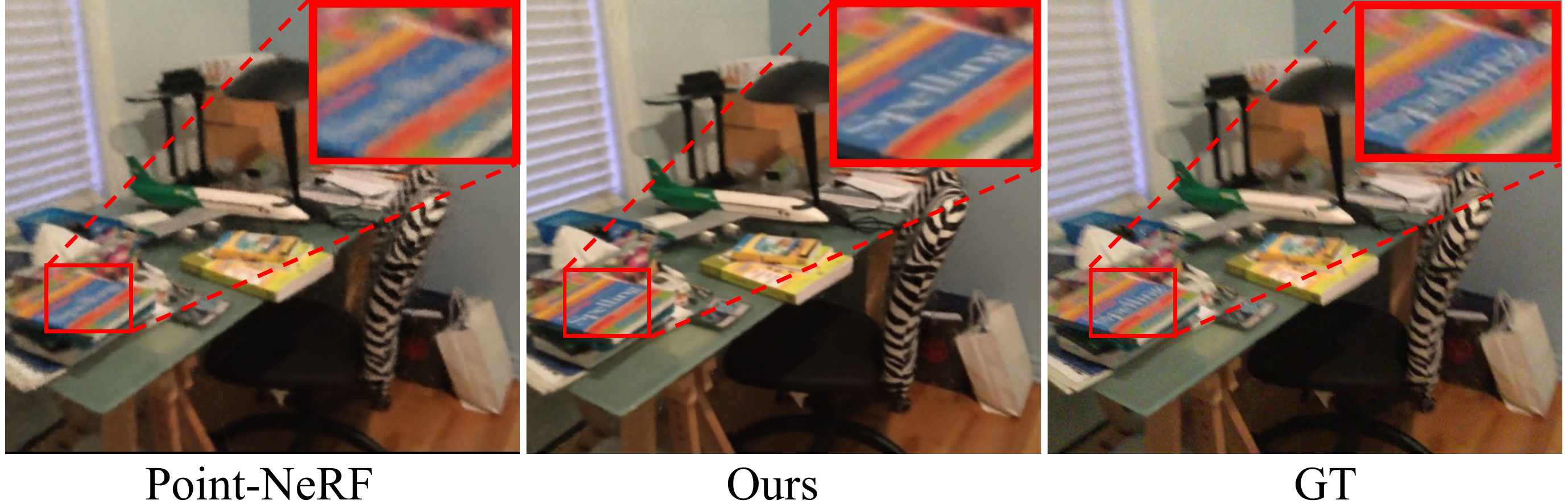}    
     \vspace{-0.25in}
     \caption{Our hybrid neural rendering model generates high-fidelity novel view images. Please note characters in the book where the result of Point-Nerf is blurry and the GT is contaminated by blur artifacts.}
    \vspace{-0.2in}
    \label{fig:teaser}
\end{figure}

Novel-view synthesis of a scene is one critical feature required by various applications, \eg, AR/VR, robotics, and video games, to name a few. Neural radiance field (NeRF) ~\cite{mildenhall2021nerf} and its follow-up works~\cite{barron2021mip, verbin2022ref, niemeyer2022regnerf, liu2020neural, yu2021pixelnerf, xu2022point} enable high-quality view synthesis on objects or synthetic data. 
However, synthesizing high-fidelity and view-consistent novel view images {of real-world large-scale scenes}
% from in-the-wild data 
remains challenging, especially in the presence of inevitable artifacts from the data-capturing process, such as motion blur (see Figure \ref{fig:teaser} \& supplementary material). 
%However, it is still challenging on large-scale real-world scenes (see Fig.~\ref{fig:teaser}) because of the limited capability of neural 3D representations and unavoidable data-capturing artifacts (\eg, blurriness due to fast motion especially on hand-held scanning devices).

%With the emerging technique of neural radiance field (NeRF)~\cite{mildenhall2021nerf},  high-quality novel-view synthesis \xjqi{of small-scale scenes} could be achieved via pre-learning a scene-specific model from a collection of posed images. However, it is still challenging to handle large-scale scenes. 
%suffering from unavoidable data-capturing artifacts (e.g. blur).

To improve novel view synthesis, mainstream research can be mainly categorized into two lines. 
% \yx{Recent efforts to improve novel view synthesis can be broadly divided into two categories.}
One line of methods directly resorts to features from training data to synthesize novel view images ~\cite{riegler2020free, wang2021ibrnet, chen2021mvsnerf, hedman2018deep}, namely image-based rendering. By directly leveraging rich high-quality features from neighboring {high-resolution }images,
% \peng{with less misalignment}
these methods have a better chance of generating high-fidelity images with distinctive details. Nevertheless, the generated images often lack consistency due to the absence of global structural regularization, and boundary image pixels often contain serious artifacts.
%\peng{and the areas around image boundaries cannot be properly handled}. 
Another line of work attempts to equip NeRF with explicit 3D representations in the form of point cloud~\cite{xu2022point, rematas2022urban}, surface mesh~\cite{yang2022neumesh, riegler2021stable} or voxel grid features~\cite{liu2020neural, fridovich2022plenoxels, yu2021plenoctrees}, namely neural 3D representation. 
Thanks to the global geometric regularization from explicit 3D representations, they can efficiently synthesize consistent novel view images but yet struggle with producing high-fidelity images in large-scale scenes (see the blurry images from Point-NeRF~\cite{xu2022point} in Fig. \ref{fig:teaser}).
This may be caused by low-resolution 3D representations~\cite{liu2020neural}, noisy geometries~\cite{aliev2020neural,dai2020neural}, imperfect camera calibrations~\cite{azinovic2022neural}, or inaccurate rendering formulas~\cite{barron2021mip}, which make encoding a large-scale scene into a global neural 3D representation non-trivial and inevitably loses high-frequency information.
% that they tend to compress a large-scale 3D scene into a relatively low-resolution 3D representation given a limited budget of memory,

%\yx{Furthermore, inaccurate depth or imperfect camera pose calibration yields noisy information, degrading the quality of rendered images.}

Albeit advancing the field, the above work all suffer immediately from low-quality training data, {\eg,} blurry images. Recently, Deblur-NeRF~\cite{ma2022deblur} aims to address the problem of blurry training data and proposed a pipeline to simulate blurs by querying multiple auxiliary rays, which, however, is computation and memory inefficient, hindering their applicability in large-scale scenes.

In this paper, we aim at synthesizing high-fidelity and view-consistent novel view images {in large-scale scenes} using in-the-wild unsatisfactory data, {\eg,} blurry data.  
First, to simultaneously address high fidelity and view consistency, we put forward a hybrid neural rendering approach that enjoys the merits of both image-based representation and neural 3D representation. Our fundamental design centers around a 3D-guided neural feature fusion module, which employs view-consistent neural 3D features to integrate high-fidelity 2D image features, resulting in a hybrid feature representation that preserves view consistency whilst simultaneously upholding quality.
% Our core design lies in a 3D-guided neural feature fusion module that effectively \peng{leverages view-consistent neural 3D features to fuse high-fidelity 2D image features} to form a hybrid feature representation that preserves view consistency without sacrificing qualities.
Besides, to avoid the optimization of the hybrid representation being biased toward one modality, we develop a random feature drop strategy to ensure that features from different modalities can all be well optimized. 

Second, to effectively train the hybrid model with unsatisfactory in-the-wild data, we design a blur simulation and detection approach to alleviate the negative impact of low-quality data on model training. 
Specifically, the blur simulation module injects blur into the rendered image to mimic the real-world blurry effects. In this way, the blurred image can be directly compared with the blurry reference image while providing blur-free supervisory signals to train the hybrid model. Besides, to further alleviate the influence of blurry images, we design a content-aware blur detection approach to robustly assess the blurriness scores of images.
% \xjqi{Besides, to further alleviate the influence of blurred images on training, we design a content-aware blur detection approach to robustly assess the blurriness score of images.}
%, which is further used to rule out the influence of blurred image contents. 
The calculated scores are further used to adjust the importance of samples during training. 
In our study, we primarily focus on the blur artifact due to its prevalence in real-world data (\eg, ScanNet); however, our ``simulate-and-detect" approach can also be applied to address other artifacts.

% We focus on the blur artifact in our study because we find it is the most common artifact in real-world data ({\eg} ScanNet), and our simulate-and-detect design is also applicable to other artifacts.  

While our model is built upon the state-of-the-art 3D- and image-based neural rendering models, our contribution falls mainly on studying their combinatorial benefits and bridging the gap between NeRF and unsatisfactory data captured in the wild. Our major contributions can be summarized as follows. 
\begin{itemize}
\item We study a hybrid neural rendering model for synthesizing high-fidelity and consistent novel view images. 
\item We design efficient blur simulation and detection strategies that facilitate offering blur-free training signals for optimizing the hybrid rendering model. 
\item Extensive experiments on real (\ie, ScanNet~\cite{dai2017scannet}) and synthetic data (\ie, Habitat-sim~\cite{habitat19iccv, straub2019replica}) showcase that our method outperforms state-of-the-art point-based methods designed for novel view synthesis. 
\end{itemize}

\begin{figure*}
    \centering
     \includegraphics[width=1.0\linewidth]{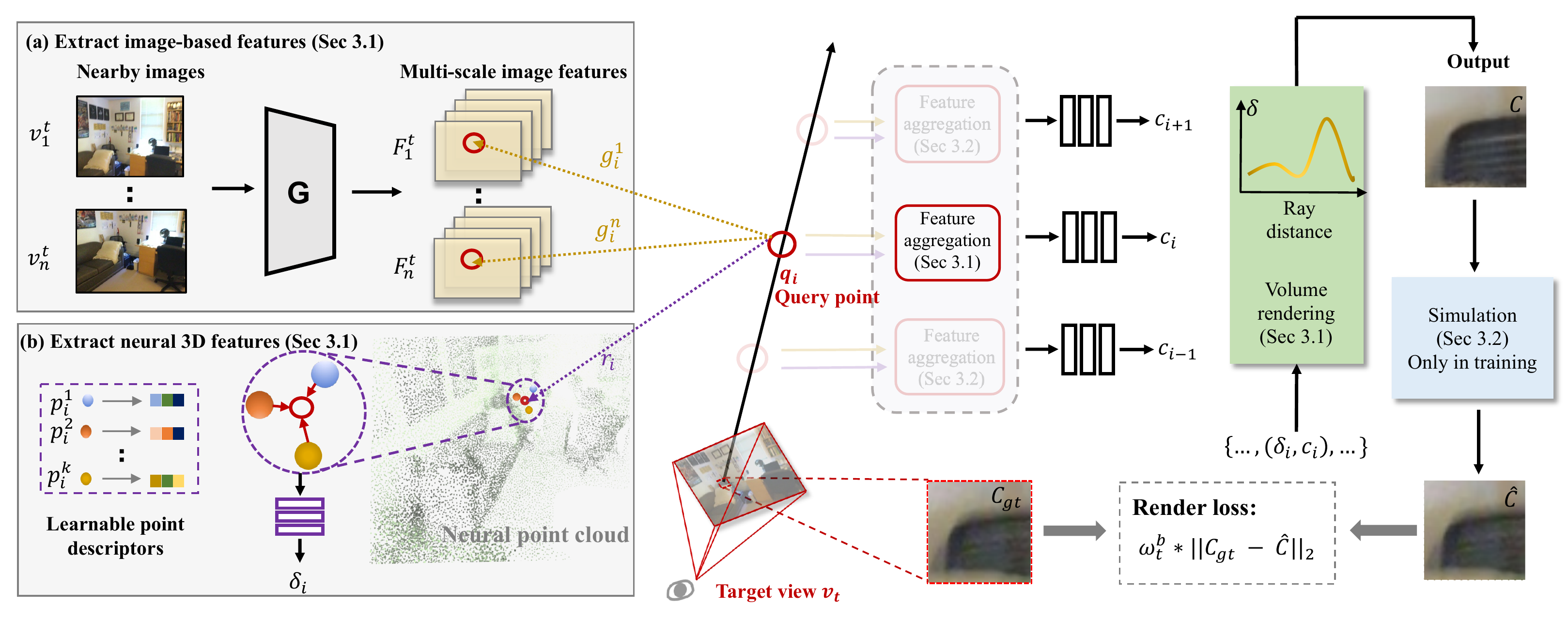}
    \caption{An overview of our hybrid neural rendering model $\mathcal{H}$. For each query point $q_i$ on a ray cast from the target view $v_t$, it has two modalities of features, \ie, (a) the image-based features $\{g_i^1, ..., g_i^n\}$ extracted from the $n$ nearby images $\{v_1^t, ..., v_n^t\}$ and (b) the neural 3D feature $r_i$ interpolated from $k$ neighboring point descriptors $\{p_i^1, ..., p_i^k\}$. To generate high-quality and consistent novel view images, our hybrid neural rendering aggregates and benefits from both features from two modalities. To handle blur artifacts in the reference images, we simulate blur effects on the rendered image patch $C$ to obtain $\hat{C}$, and then calculate the rendering loss with the ground-truth image patch $C_{gt}$. During training, we also down-weight the importance of images contaminated by artifacts according to the pre-computed quality-aware weights $\omega_t^b$ (see Sec.~\ref{section:blur}).}
    \vspace{-0.15in}
    % Specifically, the $R_i$ guides the aggregation process of $F_i^n$ with a transformer-like design and is enhanced by the aggregated image feature. Then, we predict the radiance $\overline{c}_i$ (from enhanced feature) and the volume density $\overline{\delta}_i$ (from original point descriptors) at each query point, and the output color is obtained using the volume rendering. To make both modalities well optimized, a novel random drop strategy is applied during training. Furthermore, we design a quality-aware ($\omega_i^b$) and blur simulation approach to deal with blur artifacts in training images and finally synthesize sharper novel view images.}
    \label{fig:pipeline}
\end{figure*}

\section{Related Works}

\noindent\textbf{Neural Radiance Field}
NeRF~\cite{mildenhall2021nerf} encodes the object or scene into an MLP and synthesizes novel view images through volume rendering~\cite{kajiya1984ray}. Later works extend NeRF for object manipulation~\cite{yang2022neumesh, yang2021learning, zhang2021nerfactor, jiang2022neuman} and dynamic scene modeling~\cite{park2021nerfies, park2021hypernerf, li2022neural}, etc. 
%Early attempts~\cite{barron2021mip, barron2022mip, zhang2020nerf++, verbin2022ref} to advance the design of NeRF primarily focused on theoretical perspectives. Mip-NeRF~\cite{barron2021mip}, for instance, solves the problem of varying scales by altering the way rays are cast, exhibiting a substantial improvement in rendering higher-quality images.
Recent work~\cite{zhu2022nice, xu2022point, liu2020neural} has started incorporating explicit 3D representations into NeRF training to support large-scale scenes and improve rendering details and speed. For example, Liu et al.~\cite{liu2020neural} enhance NeRF's capabilities by storing neural features in a voxel-based representation, which generates images with rich details. Similarly, Xu et al.~\cite{xu2022point} utilize a point-based neural radiance field in cooperation with point growing and pruning, which substantially speeds up training and improves the quality of the rendered image. Unlike the methods described above, we deliver a hybrid framework leveraging the advantages of neural 3D representation and image-based representation to yield high-quality images.

\vspace{0.05in}\noindent\textbf{Image-Based Rendering}
Image-based rendering is a well-known and long-standing technique~\cite{levoy1996light, debevec1996modeling} for generating novel view images. A typical pipeline is to identify a few nearby images, warp them to the target viewpoint, and then blend them to create the output~\cite{riegler2020free, hedman2018deep, hedman2016scalable}. Recently, image-based rendering methods collaborating with volume rendering have been developed for 
generalization across scenes~\cite{chen2021mvsnerf, wang2021ibrnet, yu2021pixelnerf}. For instance, IBRNet~\cite{wang2021ibrnet} employs extracted image features from neighboring images to directly predict target views without requiring per-scene optimization~\cite{mildenhall2021nerf}. 
% volume density, blending weights, and radiance value for each query point 
Since image-based rendering can directly use the rich textures from images, it typically converges faster. However, it generally suffers from temporal inconsistencies. 
% \xjqi{among neighborhood frames}
Instead, we apply the globally consistent neural 3D feature to drive the blending process in this work, improving the consistency of rendered image sequences.       

\vspace{0.05in}\noindent\textbf{Rendering with Artifacts}
For the in-the-wild environments, it is almost impossible to capture artifact-free training data due to motion blurs, noise, and environmental factors, which can adversely affect rendering quality. One solution is to restore contaminated images first~\cite{tran2021explore, xu2010two, son2021single, son2021recurrent, dai2022video, yu2022towards, wang2019edvr}, and then use restored images for training.
However, it is a challenging problem to maintain the view consistency of restored images~\cite{huang2022stylizednerf} as a pre-trained network is used to process each frame independently. 
%such two-stage schemes usually rely on pre-training and suffer from multi-view inconsistency
Recently, some works~\cite{guo2022nerfren, ruckert2022adop, ma2022deblur} have attempted to simulate the image degradation process for image restoration during training. For example, to remove reflections, Guo et al.~\cite{guo2022nerfren} propose incorporating an auxiliary MLP to model the reflection effects, which is removed during inference. 
%model the reflection and the neural radiance field using two separate MLPs, and remove the MLP modeling reflections during inference.
%\xjqi{I don't understand this sentence}. 
Rückert et al.~\cite{ruckert2022adop} propose to learn exposure-related parameters and response functions for synthesizing HDR images from training images with various exposures.
%solve the problem of images with different exposures by learning exposure-related parameters and response functions in the HDR domain. 
The work most related to us is Deblur-NeRF~\cite{ma2022deblur}, which uses auxiliary rays to simulate blurs for each training image which, however, sacrifices computation efficiency. Instead of sampling extra rays, we propose to down-weight the importance of blurry images and design a simple and efficient blur simulation method, resulting in faster training and better results.
 
\section{Method}
%\subsection{Overview}
Given RGB-D image sequences with inevitable in-the-wild artifacts,  
% \xjqi{not sure ``in-the-wild'' is a good terminology} 
our approach aims to render high-quality and consistent novel view images. 
In our study, we consider motion blur as the major artifact due to its ubiquity in data captured with hand-held devices. 
An overview of our model is shown in Fig.~\ref{fig:pipeline}. 
First, we put forward a hybrid neural rendering model $\mathcal{H}$ that incorporates neural features extracted from images and a geometry-aware neural radiance field (\eg, Point-NeRF) for producing high-quality and view-consistent synthesis results (see Sec.~\ref{section:combine}). 
Then, to produce blur-free supervisory signals for training the hybrid model, we develop a blur simulation module and a content-aware blur detection strategy to alleviate the negative impacts of blurry ground-truth reference images (see Sec.~\ref{section:blur}). 
% Specifically, the blur simulation module artificially creates blur effects within the model such that the produced blurry images $\small \hat{{C}}$ can be directly compared with the ground-truth images $\small {C}_{gt}$  for training.  
% And the blur detection strategy effectively identifies severely blurred ground-truth images and down-weight their contributions to the training objective. 
At last, we introduce the loss functions and optimization strategies for training our models (see Sec.~\ref{section:loss}).

%Our method can be divided into three parts: In Section~\ref{section:combine}, our hybrid neural rendering makes use of neural features extracted from images and the neural radiance field to render high-quality images, and applies a novel random drop strategy to make features from both modalities well-optimized; In Section~\ref{section:blur}, 
%we introduce our design to handle blur artifacts, including content-aware blur detection and blur simulation; In Section~\ref{section:loss}, it contains regularization used to guide the optimization process. And an overview of our pipeline is displayed in Fig.~\ref{fig:pipeline}.   

% Neural radiance filed can render consistent results by encoding all image information into a shared 3D representation, such as point cloud~\cite{xu2022point, aliev2020neural, dai2020neural}. On the other hand, the image-based rendering generates sharp results by directly utilizing rich textures from nearby images. Our approach takes advantages of both fields to render high-quality novel view images in . Moreover, the rendered image will be blurry when training on videos contaminated by blur artifacts. To this end, we propose a quality-aware and blur simulation design to improve the sharpness of rendered images in Section~\ref{section:blur}. At last, the regularization used to guide the optimization process is introduced in Section~\ref{section:loss}. An overview of our pipeline is displayed in Fig.~\ref{fig:pipeline}.  

% \subsection{Preliminary}
% \peng{depend on the space left}

\subsection{Hybrid Neural Rendering Model}
\label{section:combine}
Our hybrid neural rendering model is designed to combine image-based representation and the geometry-based neural radiance field for faithful and view-consistent synthesis. It consists of a neural feature extraction module to harvest information from two kinds of representations, and a neural feature fusion module to aggregate extracted neural features in a data-driven manner. Given the aggregated features, our approach renders output images based on volume rendering.  
During training, we design a random drop strategy to avoid the optimization being dominated by one of the two representations.

%Our hybrid neural rendering takes advantages of image-based rendering and \xjqi{geometry-aware} neural radiance field and consists of four steps. First, we extract neural features from both images and point-based neural radiance field. Second, the extracted features from different modalities (\ie, image features and neural radiance features) are aggregated and enhanced using a transform-like design. Third, a novel random drop strategy is applied to avoid the output being dominated by image features. At last, our approach renders output images based on the volume rendering. 
\vspace{0.05in}
\noindent\textbf{Neural Feature Extraction} 
% \xjqi{This section is rather unclear ... will modify later.... could we say sth like follow other approaches and make this section short}
As shown in Fig.~\ref{fig:pipeline}, for each query point $q_i$ on a ray cast from a target view $v_t$, we extract two modalities of features -- image-based features and neural 3D features, described as follows. 
%We will first elaborate how we extract image-based features followed by point-based features \xjqi{``neural radiance features''}. 

{\it Image-based features} (Fig.~\ref{fig:pipeline} (a)): First, we use a lightweight CNN with down-sampling layers to extract multiscale image features {\small $\{F_1^t, F_2^t, ..., F_n^t\}$} from $n$ nearby views $\{v_1^t, v_2^t,....,v_n^t\}$.  
%with images denoted as {\small $\{I_1^t, I_2^t,..., I_n^t\}$}. 
Then, the query point $q_i$ is projected to these nearby views, and features {\small $\{F_1^t(q_i), F_2^t(q_i),..., F_n^t(q_i)\}$} at the projected point location will be used to construct the image-based features for rendering. Following IBRNet~\cite{wang2021ibrnet}, we additionally add image color $v_j^t(q_i)$ and deviations of view directions $\Delta d_j^t(q_i)$ to image-based features. 
As a result, for each query point $q_i$, its image-based feature representation is $ g_i = \{g_i^1,g_i^2,...,g_i^n\}$ where $g_i^j$ is the combination of {\small $F_j^t(q_i)$}, $v_j^t(q_i)$, and $\Delta d_j^t(q_i)$. 

{\it Neural 3D features} (Fig.~\ref{fig:pipeline} (b)): We adopt a point-based neural 3D representation~\cite{xu2022point, dai2020neural, aliev2020neural} due to the wide application and high availability of point clouds. Following Point-NeRF~\cite{xu2022point}, we aggregate features from multi-view depth maps to obtain point-based 3D representations, {\ie} each point is described by a learnable descriptor. Then the neural 3D feature $r_i$ is obtained by interpolating descriptors from its $k$-nearest neighborhoods $\{p_i^1, p_i^2, ..., p_i^k\}$.
%$k$ point descriptors $\{p_i^1, p_i^2, ..., p_i^k\}$ closest to the query point $q_i$. 
Note that the point-based representation can be replaced with other geometry-based representations, such as voxel-based or mesh-based representations~\cite{liu2020neural, yang2022neumesh}.

\vspace{0.05in}
\noindent\textbf{Neural Feature Aggregation} 
\begin{figure}
    \centering
     \includegraphics[width=0.95\linewidth]{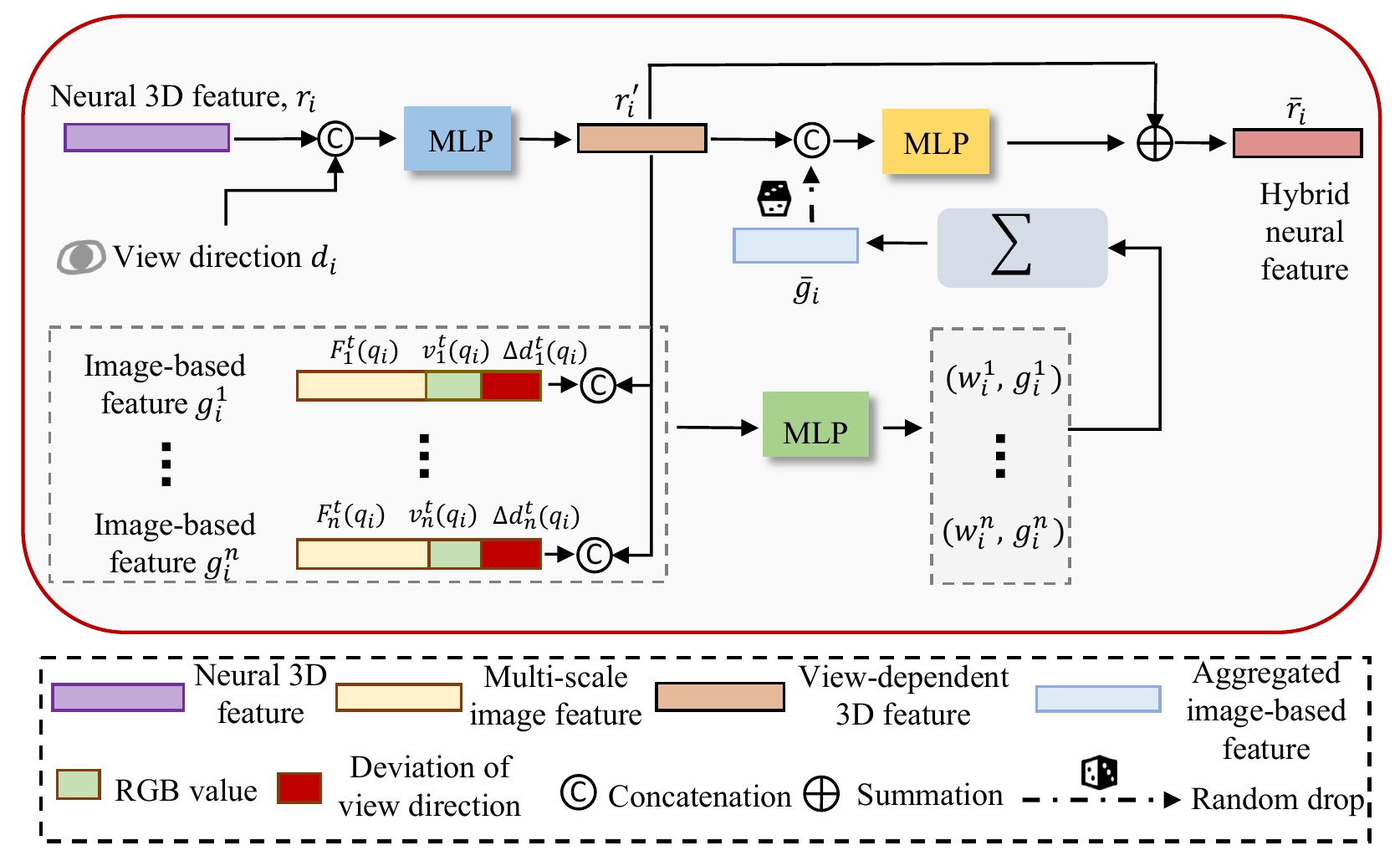}
    \caption{An overview of our feature aggregation module. The neural 3D feature $r_i$ and multiple image-based features $g_i^n$ are aggregated to generate a hybrid neural feature $\hat{r}_i$.}
    \vspace{-0.15in}
    \label{fig:feature_aggregation}
\end{figure}
As shown in Fig.~\ref{fig:feature_aggregation}, given image-based features $g_i$ from $n$ nearby views and neural 3D features $r_i$, we design a learnable method to aggregate them to form a hybrid feature $\overline{r}_i$ for each query point $q_i$, 
% which is the core component in our hybrid neural rendering model. 

First, the neural 3D feature $r_i$ combined with the view direction {$d_i$} is fed into an MLP to produce a view-dependent neural 3D feature $r'_i = \text{MLP}(r_i, d_i)$. Then, as the neural 3D feature consistently maintains global information and is free from view occlusions, we use it together with the image-based features to generate aggregation weights $\{\omega_i^1, \omega_i^2, ...\omega_i^n\}$ via an MLP layer ({\ie}, $\omega_i^j = \text{MLP}(r'_i,g_i^j)$). Further, the aggregation weights are used to combine $\{g_i^1,g_i^2,...,g_i^n\}$ to form an aggregated image feature $\overline{g}_i$ following Eq.~\eqref{eq:aggregate_img_feats}: 
% The aggregation weights serve as a selector to encourage the model to put more emphasis on using view-consistent image features and rule out errors caused by occlusions.
\begin{equation}
    \overline{g}_i =\sum_{j=1}^n (\frac{\omega_i^j}{\gamma_i} \times g_i^j),~\text{where}~\gamma_i = \sum_{j=1}^n \omega_i^j.
    \label{eq:aggregate_img_feats}
\end{equation}
Finally, we learn a residual term for $r'_i$ to get the final hybrid neural feature $\overline{r}_i$. This is achieved by enhancing the neural 3D feature $r'_i$ using the aggregated image features $\overline{g}_i$, which can be described as:
\begin{equation}
    \overline{r}_i = r'_i + \text{MLP}(r'_i, \overline{g}_i).
    \label{eq:enhance_radiance_feature}
\end{equation}

%The high-quality hybrid neural feature $\overline{r}_i$ 
% preserving \xjqi{view-consistency and high-quality}
%is further used for rendering images with volume rendering as follows. 

\vspace{0.05in}
\noindent\textbf{Volume Rendering}
% \xjqi{This section is still not clear, please fill ....}
As illustrated in Fig.~\ref{fig:pipeline}, we use $k$ nearby geometric-consistent point descriptors $\{p_i^1, p_i^2, ..., p_i^k\}$ to predict volume density $\delta_i$  considering the view-independent nature of 3D geometry. The radiance values $c_i$ are estimated through our hybrid neural features $\overline{r}_i$, which contain rich details.
% \xjqi{this section is not clear, how  you obtain volume density and how you get the color, please detail}Since the goal of leveraging image features is to enhance the appearance of rendered images, and the structure consistency should not be disturbed. Therefore, we predict the volume density from the original point descriptors ($[P_i^0, ..., P_i^k] \rightarrow [\delta_i^0, ..., \delta_i^k] \rightarrow \overline{\delta}_i$) and use the enhanced neural radiance feature to generate the RGB value ($\overline{R}_i \rightarrow \overline{c}_i$) for each query point.
Then, we apply the volume rendering~\cite{mildenhall2021nerf} to get the output color $c$ of each ray following Eq.~\ref{eq:volume_rendering}: 
\begin{equation}
\begin{aligned}
    c =& \sum_{i=1}^{M}\tau_i(1-\text{exp}(-\delta_i\Delta_i))c_i,\\
    &\tau_i = \text{exp}(-\sum_{t=1}^{i-1}\delta_t\Delta_t).
\end{aligned}
\label{eq:volume_rendering}
\end{equation}
Here, M indicates the number of query points on a ray; $\Delta_i$ represents the distance between two adjacent query points along the ray, and the $\tau_i$ means volume transmittance. 

\begin{comment}
To aggregate N nearby image features, we design a transformer-like aggregation pipeline, as illustrated in Fig.~\ref{fig:pipeline}. Specifically, the globally consistent neural radiance feature ($R_i$) is first merged with the view direction ($d$) to generate a view-related neural radiance feature ($\text{MLP}(R_i, d)\rightarrow R_i^{'}$; Query), which is then concatenated with each image feature ($F_i^n$; Key) to predict an aggregation weight ($\omega_i^n$) via a shared MLP. Following Eq.~\ref{eq:aggregate_img_feats}, we aggregate multiple image features ($F_i^n$; Value) using normalized weights.
\begin{equation}
    \overline{F}_i =\sum_{n=1}^N( \frac{\omega_i^n}{\sum_{j=1}^N \omega_i^j} \times F_i^n).
    \label{eq:aggregate_img_feats}
\end{equation}
Next, we enhance the neural radiance feature $R_i^{'}$ on the basis of the aggregated image feature ($\overline{F}_i$) by learning residuals, as described in Eq.~\ref{eq:enhance_radiance_feature}.
\begin{equation}
    \overline{R}_i = R_i^{'} + \text{MLP}(R_i^{'}, \overline{F}_i).
    \label{eq:enhance_radiance_feature}
\end{equation}
\end{comment}

% \begin{figure}
%     \centering
%      \includegraphics[width=0.95\linewidth]{images/random_drop.pdf}
%     \caption{Different random drop strategies. The left one indicates the ray-based random drop. The right one indicates the query-point-based random drop. \xjqi{figur need to be updated}}
%     \label{fig:random_drop}
% \end{figure}
\vspace{0.05in}
\noindent\textbf{Random Drop} We develop two random drop strategies that randomly drop image features during optimization to ensure both modalities of features can be well-optimized: 1) the {\it ray-based random drop} will drop all image features on randomly selected rays; 2) the {\it query-point-based random drop} will randomly select query points on all rays and then remove all image features on them. The motivation behind the random drop is that we find the optimization of the hybrid representation can be easily dominated by image features, leaving neural 3D features poorly trained. This is because the image features are very similar to the reference images and are thus more easily optimized. Unless otherwise specified, we adopt the ray-based random drop during training. 
In the experiment part (see Fig.~\ref{fig:Random_drop_results}), we show the effects of the two strategies.

% To make both modalities of features well-optimized during training, we develop random drop strategies that randomly remove image features from query points to force neural radiance features to participant in the optimization process. In this paper, we design two different random drop strategies. 1) \textbf{Ray-based random drop}. In Fig.~\ref{fig:random_drop} left, we randomly select rays, and then the image features of all query points on these rays are removed.  2) \textbf{Query-point-based random drop}. In Fig.~\ref{fig:random_drop} right, we  By doing this, the network is not bias towards one modality (\ie, image features) that is closer to the reference and is easier to be optimized. 

% Consequently, image features will dominate the output, resulting in unstable results and poor image quality in regions not covered by image features (Fig.~\ref{fig:No_random_drop}). 

\subsection{Blur Simulation and Detection}
\label{section:blur}
%\xjqi{please state sth about the image captured in the wild}
%\xjqi{The captured posed images in the wild for training suffer from unavoidable motion blur, which will result in blurry rendering regardless of the rendering model. To address this issue, we propose xxx }
We propose two complementary strategies to address the negative influence of blurry reference images on optimizing the hybrid neural rendering model. First, we design a simulation method that simulates blur effects on the rendered image patch $C$ to imitate the blur effects of the reference image patch $C_\text{gt}$. 
By comparing the blurred image patch $\hat{C}$ with the reference image patch during training, the sharpness of the rendered images can be preserved. 
Second, we develop a content-aware detection method to pre-compute the blurriness scores of reference images and down-weight the importance of blurry images based on the calculated scores. 
The two strategies work collectively to address the data quality challenge. 
%blurriness scores from a content-aware blur detection method. 
%To avoid the optimization of the hybrid neural rendering model \xjqi{replace to notation if we have} being influenced by blurry reference images
%we design blur simulation, injecting blur effects on rendered image \xjqi{$\mathcal{I}$} during training, and blur detection, estimating the degree of blurriness of ground-truth images to down weight the contributions of severely blurred ones to the training objective.
%To improve the sharpness of rendered images, we propose to handle blur artifacts in the training images from two perspectives. First, we propose to down weight the importance of blurry images based on a content-aware blur detection. Second, we reduce the influence of blur artifacts by simulating blurs during training.    

\vspace{0.05in}
\noindent\textbf{Blur Simulation}
\begin{figure}
    \centering
     \includegraphics[width=1.0\linewidth]{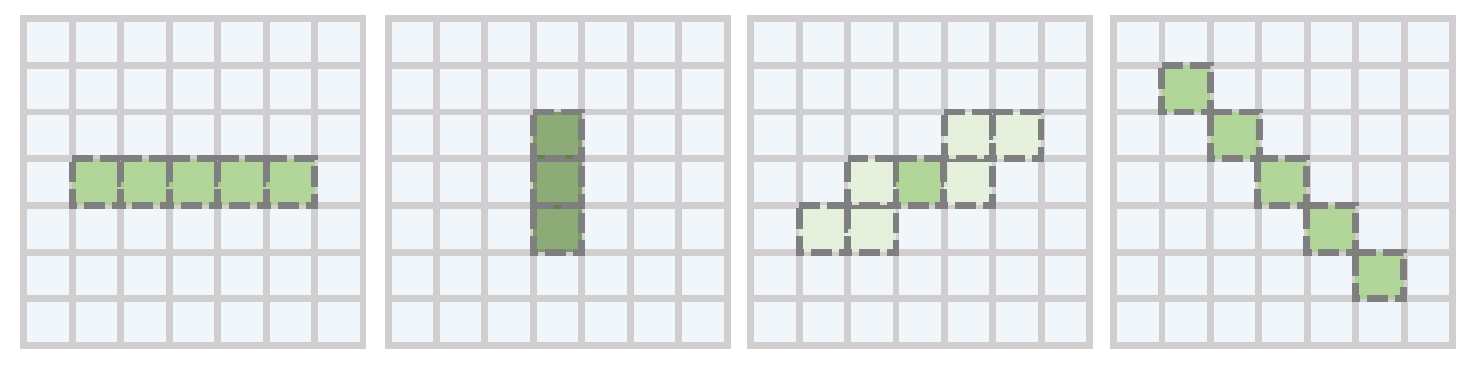}
    \caption{Examples of blur kernels. The pre-defined blur kernels have different moving directions and distances.}
    \vspace{-0.15in}
    \label{fig:blur_kernel}
\end{figure}
% The down-weighting operation does not completely exempt our method from blur artifacts and may result in fewer images. 
%We simulate blurs at the same time to decrease the influence of blurry images. Specifically, 
% \xjqi{please refine to add more insights}
% We simulate blur effects within the model in the hope of approximating blur effects in the image-capturing process, so that the blurred output $\hat{C}$ can be directly compared with the blurry reference image $C_\text{gt}$ and provides blur-free supervisory signals to train the hybrid rendering model. 
% \xjqi{We assume that the camera is moving in one direction with a certain distance at a time. Thus, xxx blur kernel is adopted }
%Our blur simulation assumes that the camera moves in one direction through a certain distance while capturing high frame-rate videos, which incurs motion blurs. 
To simulate motion blur, we assume that the camera moves in one direction through a certain distance while capturing high frame rate videos. 
Specifically, we take into account $N_v$ directions and $N_d$ distances for creating blur kernels ($B_i|i=0, 1, ..., N_v\times N_d$) that are used to simulate blurs, and some examples of blur kernels are shown in Fig.~\ref{fig:blur_kernel}. When $i=0$, it means no blur simulation.
To determine which blur kernel approximates the blur effects best, we first apply all blur kernels to the rendered results to obtain the blurred image patches $\hat{C}_{i} = \text{Conv}(C, B_i)$, and then choose the blur kernel $i$ that yields an output patch $\hat{C}_{i}$ with the minimum photo-metric loss w.r.t the reference image patch $C_\text{gt}$. This process is described as:
\begin{equation}
\begin{aligned}
    &L_{i} = ||\hat{C}_{i} - C_\text{gt}||_2, \\ 
    \mathcal{L}_\text{color} = &\min\{L_{i}|i=0,1,...,N_v\times N_d\}.
\end{aligned}
\label{eq:color_loss}
\end{equation}
Because our blur simulation does not need to render extra rays nor exhaustively optimize per-view embeddings as in Deblur-NeRF~\cite{ma2022deblur}, it runs faster and is more memory efficient, making it suitable for large-scale scenes with dense views. This blur simulation process is removed during inference to produce sharp images $C$. (In appendix~\ref{sec: Extendable}, we further propose a novel learnable scheme to efficiently predict degradation kernels by exploring differences between rendered and reference patches.)
% \xjqi{Although the blur simulation cannot exactly approximate xxx, [explain why it helps to train a better rendering model]}

\begin{figure}
    \centering
     \includegraphics[width=1.0\linewidth]{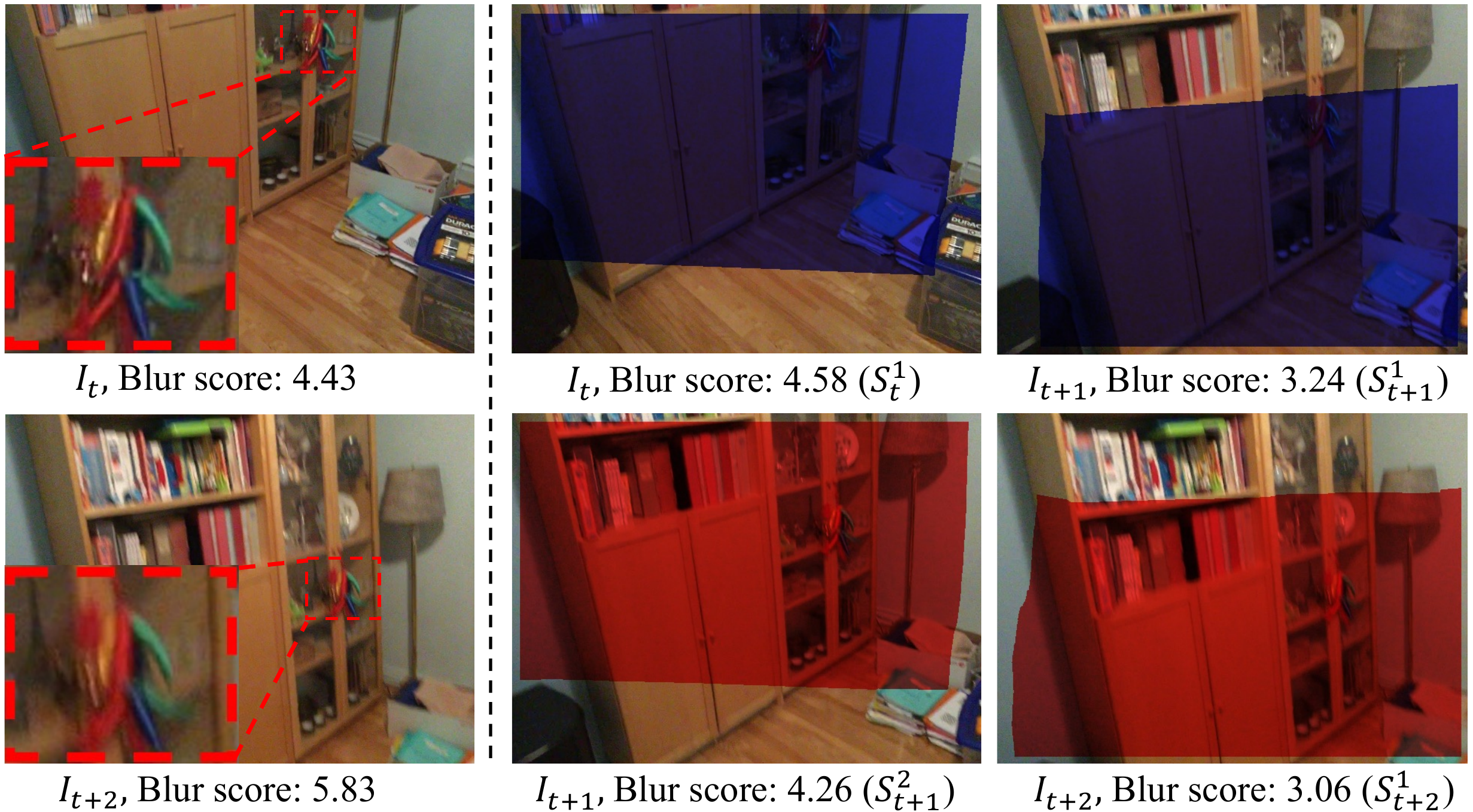}
     \vspace{-0.2in}
    \caption{Content-aware blur detection. Left: the blurriness score~\cite{pech2000diatom} (large is sharper) is highly affected by image contents, and is usually low when the image contains textureless contents (\eg, floor).  Right: the content-aware blur detection computes blurriness scores on overlapping regions of two images, thus obtaining more accurate scores.}
    \vspace{-0.15in}
    \label{fig:content_aware}
\end{figure}

\vspace{0.05in}
\noindent\textbf{Content-Aware Blur Detection} 
In addition, we also down-weight the contribution of blurry images based on the blurriness score (a smaller value indicates more severe blur artifacts). However, we find that the ``variation of the Laplacian"~\cite{pech2000diatom} method used to compute blurriness scores is prone to be influenced by image contents, thus unsuitable for scoring the reference images directly. As shown in the left of Fig.~\ref{fig:content_aware}, the upper image is sharper than the bottom one but has a lower blurriness score. This is because the upper image contains more textureless contents (\ie, the floor).

To exclude the influence of image contents, we develop a content-aware blur detection approach, which outputs accurate blurriness scores by scoring the overlapping regions. As shown in Fig.~\ref{fig:content_aware} right, our method first takes two neighboring images $\{I_t, I_{t+1}\}$ as inputs and estimates their overlapping regions (blue areas in Fig.~\ref{fig:content_aware}) using optical flow~\cite{teed2020raft}. Then, it returns two images' blurriness scores $\{S_{t}^1, S_{t+1}^1\}$ calculated from the overlapping regions.
% ~\cite{pech2000diatom} 
Next, to compute the blurriness score of image $I_{t+2}$, we use another image pair $\{I_{t+1}, I_{t+2}\}$ and repeat the process above to obtain two new blurriness scores $\{S_{t+1}^2, S_{t+2}^1\}$. Considering different overlapping regions in an image (\eg, blue and red regions of $I_{t+1}$ in Fig.~\ref{fig:content_aware}) will lead to different blurriness scores $S_{t+1}^1$ and $S_{t+1}^2$, we align them by scaling $S_{t+1}^2$ to $S_{t+1}^1$. Correspondingly, the blurriness score of $I_{t+2}$ is scaled following $S_{t+2}^1 = S_{t+1}^1/S_{t+1}^2\times S_{t+2}^1$. Similarly, the blurriness scores of other images can be computed. 
Please refer to the supplementary file for details. 
%\peng{(Please check the supplementary file for details)} 
Finally, we convert blurriness scores into quality-aware  weights $\omega_t^b$ following:
\begin{equation}
    \omega_t^b = (\frac{N\times S_t^1}{\sum_{t=0}^{N }S_t^1})^{\alpha},
    \label{eq:blur_weight}
\end{equation}
where $N$ represents the number of images, and $\alpha \ge 0$ is a hyper-parameter used to adjust the distribution of quality-aware image weights. These weights are further applied to the training objective in Sec.~\ref{section:loss} to down-weight the importance of blurry images. Alternatively, you can use $\omega_t^b$ as sampling probabilities to sample training images.

\subsection{Optimization}
\label{section:loss}
% \xjqi{please also include these in your figure for better illustrations}
% Following Point-NeRF, we employ the same training regularization and strategies (\ie, point growing and pruning), and merge our design for handling blur artifacts (see Sec.\ref{section:blur}) into the training regularization. 
Our training objective consists of a photometric loss $\mathcal{L}_\text{color}$ in Eq.~\eqref{eq:color_loss} that requires the rendered image patch $C$ after blur simulation $\hat{C}$ to have the same appearance as the reference image patch $C_\text{gt}$; and a sparsity loss $\mathcal{L}_\text{sparse}$~\cite{lombardi2019neural, xu2022point} that encourages each point to have a confidence of 0 or 1 for the follow-up point pruning and growing operations. 
Following Point-NeRF~\cite{xu2022point}, the point growing and pruning operations are applied every 10k iterations. 
%based on the confidence of points, the predicted volume density of each query point, and the point density of neighborhood points.  
% \xjqi{I think the blur simulation is incorporated in your final images right?? the following is not clear how you calculate $L_color$ from the neural rendering model or after blur simulation, please organize them  as a whole}
After incorporating the quality-aware design ($\omega_t^b$ in Sec.~\ref{section:blur}), the final training objective is defined as:
\begin{equation}
    \mathcal{L}_t = \omega_t^b(\mathcal{L}_\text{color} + \beta \mathcal{L}_\text{sparse}),
    \label{eq:loss}
\end{equation}
where $\beta = 0.002$ is used to balance different loss terms and $\omega_t^b$ is the estimated blurriness score to down weight blurry images (see Section \ref{section:blur}).

% And point descriptors $P$ and network parameters, including the image feature extractor $G$ and all MLPs in the hybrid neural rendering model $\mathcal{H}$, are optimized.   

\section{Experiments}
\subsection{Implementation Details}
\noindent\textbf{Network and Training} 
The 2D CNN ($G$) used to extract image features has three down-sampling layers, and the point-based neural 3D representation is constructed following Point-NeRF~\cite{dai2017scannet}. We select four neighboring frames ($n=4$) and eight nearest point descriptors ($k=8$) to extract neural features. We train our models using the Adam~\cite{kingma2014adam} optimizer with an initial learning rate of $0.0005$. 
A total of 200k iterations are used for training. 
%After optimizing the model by 200k iterations for each scene, the results are reported.

\vspace{0.05in}
\noindent\textbf{Blur Simulation}
We build our blur kernels considering $N_v=4+8$ directions (\ie, `left-right', `up-down', `top left-bottom down', and `bottom left-top right'; both symmetrical and asymmetrical) and three moving distances $N_d=3$ (\ie, 1, 2, 4). 
% \peng{Convolving with asymmetric kernels will shift pixels, which somewhat mitigates misalignment between rendered and reference patches.} 
To apply blur simulation, we sample $8\times8$ patches with dilations~\cite{schwarz2020graf} during the training, and the $\alpha$ in Eq.~\eqref{eq:blur_weight} is set as 1.

\vspace{0.05in}
\noindent\textbf{Dataset} 
We conduct our experiments on ScanNet~\cite{dai2017scannet} and synthetic data generated from Habitat-sim~\cite{habitat19iccv}. 
1) ScanNet~\cite{dai2017scannet} contains RGB-D image sequences captured in large-scale indoor scenes with handheld sensors. Following Point-NeRF, we conduct experiments on ``Scene0101\_04'' and ``Scene0241\_01'' and select every fifth image for training and the remaining images for testing. Note that images in the ScanNet are blurry, which is not suitable for quantitatively evaluating the sharpness of rendered images. Thus, we additionally evaluate our method using synthetic data. 2) Habitat-sim is a simulator~\cite{habitat19iccv, szot2021habitat} that synthesizes blur-free RGB-D sequences of large-scale scenes (\ie, 'VangoRoom' and 'LivingRoom'~\cite{straub2019replica}). We then add motion blurs to the synthesized training sets. Please see the supplementary file for details. 
%and then only add blurs~\cite{LeviBorodenko20} to the training split every fifth image, and the remaining blur-free test split is used to obtain quantitative results.

\vspace{0.05in}
\noindent\textbf{Baselines} We compare our method with other representative image-based and neural-radiance-based novel view synthesis approaches, including: 1) NeRF~\cite{mildenhall2021nerf}; 2) IBRNet~\cite{wang2021ibrnet} which combines image-based rendering with volume rendering and generates high-quality novel view images without using depth; 3) NPBG~\cite{aliev2020neural} which renders images using a U-Net-like design by rasterizing point descriptors onto the image plane
%, NPBG renders images using a U-Net-like design. 
4) Point-NeRF~\cite{xu2022point}, which is the state-of-the-art point-based method for novel view synthesis combining point-based neural representation and neural radiance field with volume rendering; and 5) Deblur-NeRF~\cite{ma2022deblur} which improves the sharpness of rendered images by simulating the blurring process with a deformable sparse kernel module.

\subsection{Results on ScanNet}
\label{sec:scannet}
Quantitative comparisons with other baselines in terms of PSNR, SSIM, and LPIPS~\cite{zhang2018unreasonable} are reported in Table~\ref{tab:ScanNet}. Our hybrid neural rendering design ``Ours (H)" outperforms previous methods by enhancing the quality of neural 3D representations. However, the PSNR and SSIM drop in the full version of our method ``Ours". 
%This is because the ScanNet data contains heavy blurriness (See Fig.~\ref{fig:ScanNet} last column), which can adversely affect the metrics and is unsuitable for quantitative evaluation of the blur-handling module that changes the sharpness distribution of rendered images.
This is because our blur-handling modules mimic blurriness effects and down weight blur images, enabling the model to learn from clean supervision. However, since this differs from the original training data distribution, the model may not fit the evaluation metric well. 
Moreover, Deblur-NeRF delivers a low PSNR because it tends to introduce misalignment between rendered and reference images.   

\begin{figure*}
    \centering
    \includegraphics[width=1.0\linewidth]{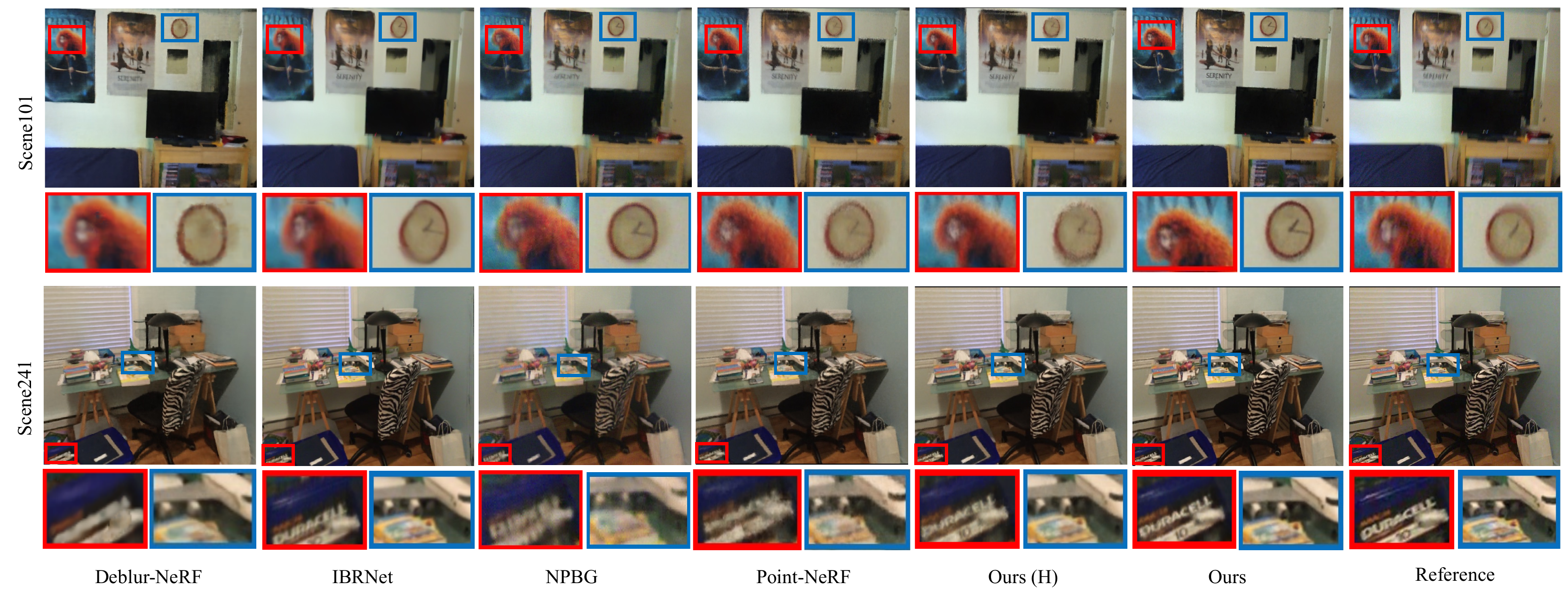}
    % \vspace{-0.25in}
    \caption{Qualitative comparisons on ScanNet. The highlighted regions are zoomed-in and placed at the bottom for better comparisons. From the results, our method can synthesize sharper images than other approaches that are suffering from blurriness, distortions, and jagged edges. Moreover, the sharpness is further improved after applying our design to handle blur artifacts (\ie, Ours vs. {Ours (H)}).}
    \vspace{-0.0in}
    \label{fig:ScanNet}
\end{figure*}

We show qualitative comparisons in Fig.~\ref{fig:ScanNet}. Our method can render high-quality novel view images while other baselines suffer severely from blurriness and distortions. For example, the clock on the wall is distorted with IBRNet, and the book generated by Point-NeRF is blurry. {In contrast, our model produces results with clear characters in the book (Fig.~\ref{fig:ScanNet} ``Ours (H)"), validating the efficacy of our hybrid representation.}
%the text in the book is clear, and the clock looks natural in our method. 
%We attribute these good performances to making use of both image-based features and neural 3D features. 
%The high-quality results validate the efficacy of our hybrid representation.
Further, the rendered images become sharper when using our design to handle blur artifacts; please notice the human face on the poster (Fig.~\ref{fig:ScanNet} ``Ours"). To better demonstrate the efficacy of our approach in rendering consistent results, we provide videos in the supplementary file: {our results are more temporally consistent than the image-based rendering (\ie, IBRNet), thanks to the globally consistent neural 3D features.}
% As for the temporal consistency of synthesized videos, our results are more consistent (thanks to the globally consistent neural 3D features) than the image-based rendering method, \ie, IBRNet (please refer to the video results in the supplementary material).            

\begin{table}[]
    \centering
    \resizebox{1.0\columnwidth}{!}{
    \begin{tabular}{c|ccc|ccc}
    \hline
     & \multicolumn{3}{c}{Scene101\_04} & \multicolumn{3}{|c}{Scene241\_01}\\
    & PSNR$\uparrow$ & SSIM$\uparrow$ & LPIPS$\downarrow$ & PSNR$\uparrow$ & SSIM$\uparrow$ & LPIPS$\downarrow$\\
    \hline
    Point-NeRF~\cite{xu2022point}& 29.88 & 0.913 & 0.203 & 30.54 & 0.910 & 0.236\\
    IBRNet~\cite{wang2021ibrnet}&  29.55 & 0.811 & 0.307&  21.49 &  0.755 & 0.368\\
    NPBG~\cite{aliev2020neural}& 26.33 & 0.871 & 0.187 & 27.34 & 0.841 & \textbf{0.188} \\
    Deblur-NeRF~\cite{ma2022deblur} & 24.55 & 0.693 & 0.308 & 20.66 & 0.652 & 0.401\\
    NeRF~\cite{mildenhall2021nerf} & 27.16 & 0.730 & 0.350 &  21.69& 0.610 & 0.494\\
    \hdashline
    Ours (H) & \textbf{30.33} & \textbf{0.919} & 0.186 & \textbf{31.25} & \textbf{0.918} & 0.218\\
    Ours & 29.33 & 0.909 & \textbf{0.181} & 30.78 & 0.914 & 0.206\\
    \hline
    \end{tabular}}
    \caption{
    Quantitative comparisons on ScanNet. ``Ours (H)": use hybrid neural rendering without handling blur artifacts. We use PSNR, SSIM, and LPIPS to evaluate the rendering quality ($\downarrow$: small is better; $\uparrow$: large is better). Our method outperforms all other baselines by a large margin, especially on PSNR. Note that the full version of our method (``Ours") is worse on the PSNR and SSIM, this is because the reference images in ScanNet are blurry.} 
    \vspace{-0.1in}
    \label{tab:ScanNet}
\end{table}

\subsection{Results on Synthetic Data}
\begin{figure*}
    \centering
    \includegraphics[width=1.0\linewidth]{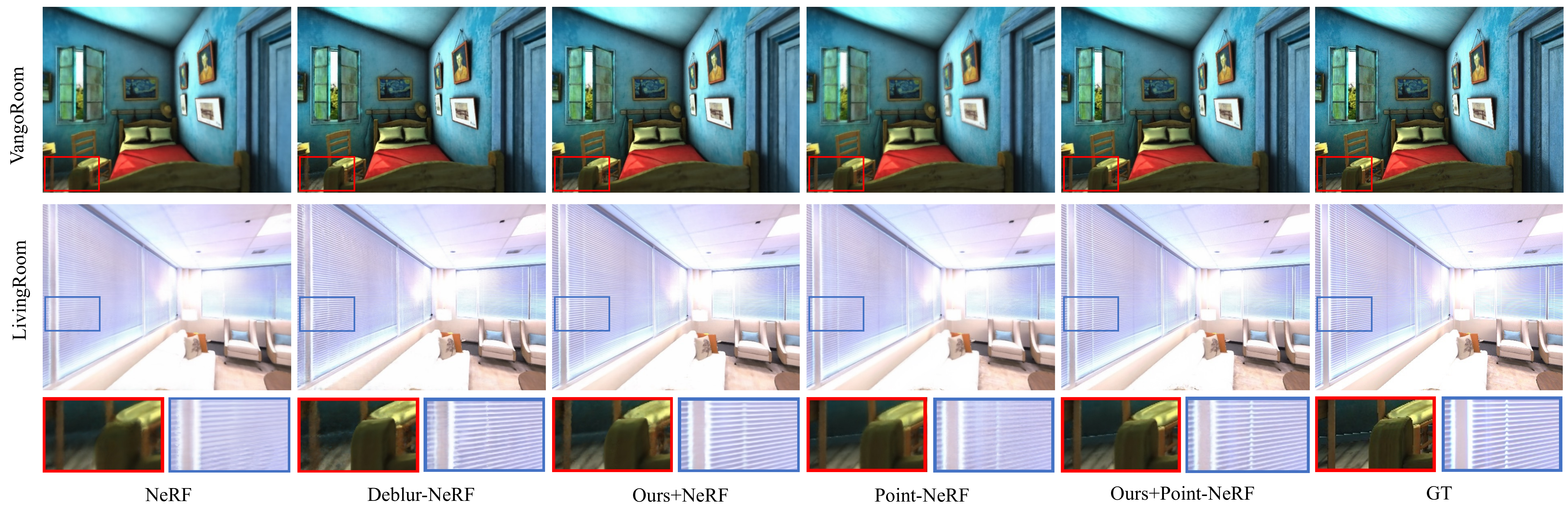}
    % \vspace{-0.25in}
    \caption{Qualitative comparisons on the synthetic data. We validate our designs to deal with blur artifacts on two different frameworks, \ie, NeRF and Point-NeRF. By applying our design, the details of rendered images become sharper. Besides, the Deblur-NeRF can also improve the sharpness, but the structure (\eg, blinds) is distorted.}
    \label{fig:Blurry_synthetic}
    \vspace{-0.0in}
\end{figure*}

We conduct experiments on the synthetic data to validate our designs to handle blurriness. 
In particular, we incorporate our designs into two different frameworks (\ie, NeRF and Point-NeRF) to show its generalization ability. Here, we remove the image-based rendering branch on the NeRF-based framework for fair comparisons.
Fig.~\ref{fig:Blurry_synthetic} shows that our method significantly enhances the sharpness of rendered images compared to NeRF and Point-NeRF, which are also confirmed in Table~\ref{tab:Blurry_synthetic}.
Notably, images from Deblur-NeRF contain more details than NeRF but suffer from distorted image structures, such as the blinds and the table leg. 
This is possible because the learning of ray deformation is under-constrained with too many degrees of freedom and thus prone to corrupting original structures, or the view embeddings are not well-optimized due to the increased number of training views.
Our easy-to-plug-in method outperforms Deblur-NeRF on PSNR and SSIM and delivers competitive performance on LPIPS.
{It is worth noting that to achieve the above results, NeRF takes 4.5 hours, while our method takes 4.6 hours. Thus, the increase in training time brought by blur simulation is negligible. However, Deblur-NeRF needs 8.5 hours which incurs much more overheads. The time is reported with training the model on a single NVIDIA 3090 GPU. }
% (check the time again) Moreover, our method requires $25\%$ less memory consumption than Deblur-NeRF. (remove this?)}

% Importantly, to achieve the results above, our method is $1.7\times$ faster and requires only $75\%$ memory consumption of that in Deblur-NeRF. 

%From the results in Fig.~\ref{fig:Blurry_synthetic} and Table~\ref{tab:Blurry_synthetic}, our method significantly enhances the sharpness of rendered images than NeRF and Point-NeRF, and the values of all three metrics are improved on both NeRF-based and Point-NeRF-based frameworks. And images from Deblur-NeRF contain more details than NeRF, but their structures are distorted, such as the blinds and the table leg. One possible explanation is that the highly free ray deformation in Deblur-NeRF may cause distortions. Besides, our method outperforms Deblur-NeRF on PSNR and SSIM and has competitive performance on LPIPS. More importantly, to achieve the results above, our method is $1.7\times$ faster and only takes $75\%$ memory consumption of that in Deblur-NeRF.    

\begin{table}[h]
    \centering
    \resizebox{1.0\columnwidth}{!}{
    \begin{tabular}{c|ccc|ccc}
    \hline
     & \multicolumn{3}{c}{VangoRoom} & \multicolumn{3}{|c}{LivingRoom}\\
     & PSNR$\uparrow$ & SSIM$\uparrow$ & LPIPS$\downarrow$ & PSNR$\uparrow$ & SSIM$\uparrow$ & LPIPS$\downarrow$\\
     \hline
    NeRF & 28.83 & 0.769 & 0.339 & 29.73 & 0.848 & 0.215  \\
    Deblur-NeRF~\cite{ma2022deblur} & 29.30 & 0.793 & \textbf{0.247} & 31.82 & 0.895 & 0.132\\
    Ours+NeRF & \textbf{30.26} & \textbf{0.805} & 0.259 & \textbf{32.70} & \textbf{0.912} & \textbf{0.124}\\
    \hdashline
    Point-NeRF & 31.24 & 0.950 & 0.152 & 32.20 & 0.959 & 0.109 \\
    Ours+Point-NeRF & \textbf{33.27} & \textbf{0.966} & \textbf{0.097} & \textbf{35.30} & \textbf{0.980} & \textbf{0.051}\\
    % \hdashline
    % Deblur-Image (hold on)~\cite{} &  & & &  &  & \\
    \hline
    \end{tabular}}
    \caption{Quantitative comparisons on the synthetic data. We apply our design used to handle blur artifacts to two different frameworks, \ie, NeRF and Point-NeRF. With our design, the values of all three metrics receive significant improvements.} 
    \vspace{-0.1in}
    \label{tab:Blurry_synthetic}
\end{table}

\subsection{Ablation Studies}

In this section, we conduct comprehensive ablations of the proposed designs in our method.

\vspace{0.05in}
\noindent\textbf{Advantages of Image Features}
We first assess the contribution of image features in our system by comparing ``Ours (H)" and Point-NeRF (\ie, without using image features). 
%We compare different methods with (\ie, ``Ours (H)") and without (\ie, Point-NeRF) using image features.
As shown in Fig.~\ref{fig:ScanNet} and Table~\ref{tab:ScanNet}, our method benefits from image features and outperforms Point-NeRF.     
{Moreover, our hybrid model converges faster. For example, we achieve PSNR 31.0 after 20k iterations (80 minutes) on ``Scene241$\_$01", whereas Point-NeRF delivers 29.3 PSNR after 40k iterations (84 minutes). (Please refer to supplementary material for more results.)} This is because compressing all information into a neural 3D representation is difficult since it requires accurate camera poses, high-resolution 3D representations, etc. On the contrary, high-fidelity image features can directly compensate for defective neural 3D features and enable synthesizing high-quality results with  fewer training iterations.  

% \begin{figure}
%     \centering
%     \includegraphics[width=0.95\linewidth]{images/Ab_image_features.pdf}
%     \caption{Advantages of using image features. This figure plots the trend of PSNR at different training iterations. When using image features (\ie, hybrid neural rendering), the performance is better than the neural-3D-feature-only design (\ie, Point-NeRF).\peng{move to the supp}} 
%     \label{fig:Ab_imageFeats}
%     \vspace{-0.1in}
% \end{figure}

\begin{figure}
    \centering
    \includegraphics[width=1.0\linewidth]{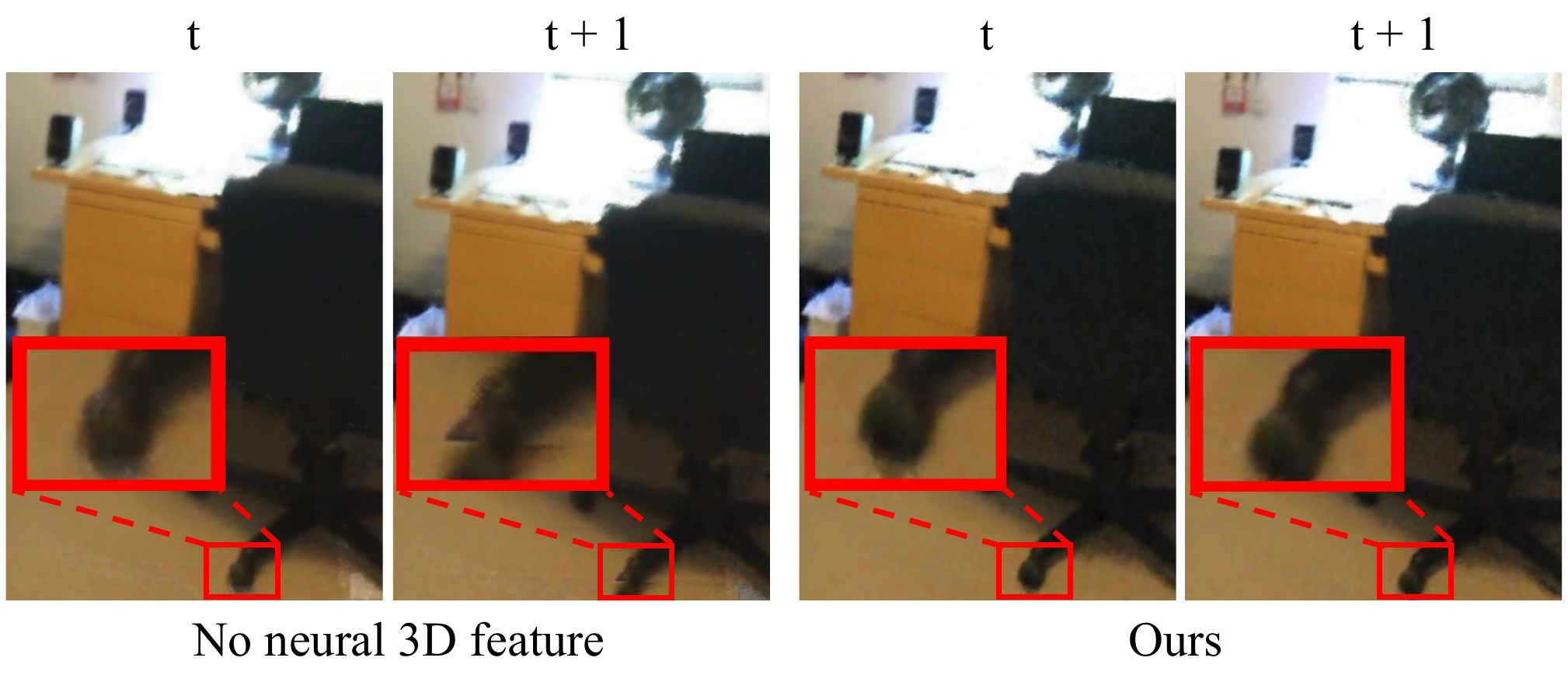}
    \caption{Advantages of neural 3D features. Without using the neural 3D feature, the rendered images will be inconsistent at different times. This can be observed in the example of the chair leg.}
    \label{fig:Ab_globalFeats}
    \vspace{-0.1in}
\end{figure}

\vspace{0.05in}\noindent\textbf{Advantages of Neural 3D Features}
We then show the value of the neural 3D feature by comparing it with IBRNet, which uses only image features.
%We first compare our method with an image-based approach (\ie, IBRNet). 
From the results in Fig.~\ref{fig:ScanNet} and the video in the supplementary material, the rendered images from IBRNet are often distorted and inconsistent due to the lack of global 3D constraints. To further investigate the efficacy of learned neural 3D representations, we replace the neural 3D features with the mean and variance of image features extracted from nearby frames in the feature aggregation module (see Fig.~\ref{fig:feature_aggregation}). 
The corresponding results are displayed in Fig.~\ref{fig:Ab_globalFeats}, our approach preserves the shape consistency of the chair leg. 
% which has advantages over the method without using neural 3D features.

\vspace{0.05in}\noindent\textbf{Blur Simulation and Quality-aware Design}
We further show the effect of blur simulation and detection by removing each of them at a time, and the results on ``LivingRoom" are shown in Table~\ref{tab:Ab_blurSim}.
%We validate the effectiveness of each component designed to handle blur artifacts on the ``LivingRoom". 
According to Table~\ref{tab:Ab_blurSim}, both components contribute to the final performance, and the performance is improved when they are combined.
%One explanation is that the quality-aware design down-weight the severe blur that is difficult for blur simulation to handle. On the contrary, blur simulation mitigates the influence of blur artifacts remaining in the image pool after applying quality-aware design.         

\begin{table}[h]
    \centering
    \resizebox{1.0\columnwidth}{!}{
    \begin{tabular}{c|ccc|ccc}
    \hline
     & \multicolumn{3}{c}{NeRF-Based} & \multicolumn{3}{|c}{Point-NeRF-Based}\\
     & PSNR$\uparrow$ & SSIM$\uparrow$ & LPIPS$\downarrow$ & PSNR$\uparrow$ & SSIM$\uparrow$ & LPIPS$\downarrow$\\
    \hline
    Baseline & 29.73 & 0.848 & 0.215 & 32.20 & 0.959 & 0.109\\
    + Blur simulation & 32.24 & 0.905 & 0.133 & 34.32 & 0.974 & 0.065\\
    + Quality-aware weight & 31.81 & 0.900 & 0.148 & 34.48 & 0.977 & 0.065\\
    Full & 32.70 & 0.912 & 0.124 & 35.30 & 0.980 & 0.051\\
    \hline
    \end{tabular}}
    \caption{We validate our designs to handle blur artifacts on the 'LivingRoom'. Each component improves the performance and works better when combined.}
    \vspace{-0.1in}
    \label{tab:Ab_blurSim}
\end{table}

% \begin{figure}
%     \centering
%     \includegraphics[width=1.0\linewidth]{images/blur_sim.pdf}
%     \caption{The components of blur sim on both ScanNet and blurry synthetics.}
%     \label{fig:Ab_blurSim}
% \end{figure}

\begin{figure}
    \centering
    \includegraphics[width=1.0\linewidth]{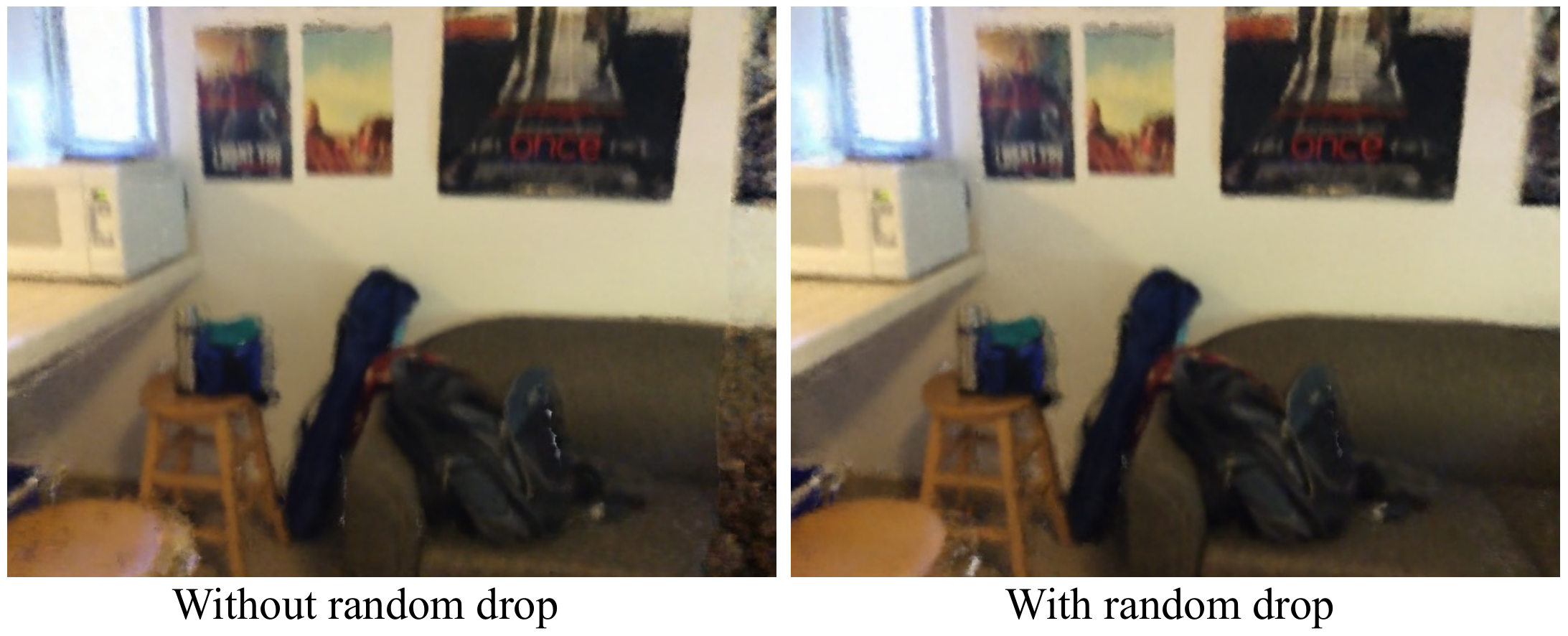}
    \caption{Efficacy of random drop. Without random drop, areas around image boundary (\eg, the sofa in the left image) not covered by image features are bad.}
    \label{fig:No_random_drop}
    \vspace{-0.in}
\end{figure}

\begin{figure}
    \centering
    \includegraphics[width=1.0\linewidth]{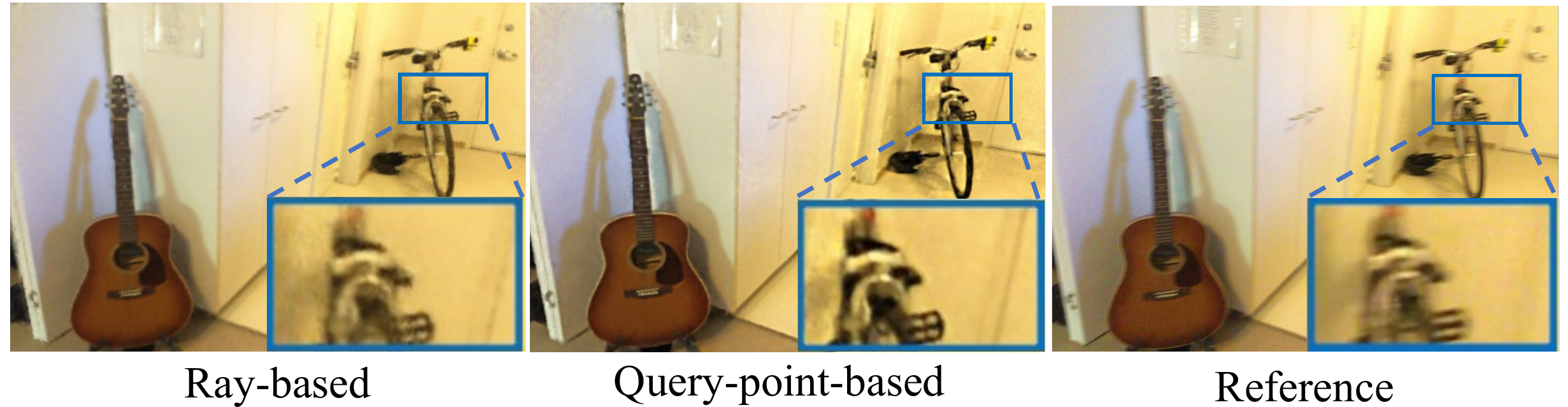}
    \caption{Different random drop methods. Query-point-based random drop automatically enhances the rendered images, but it tends to change the color tone. Please note the bicycle.}
    \vspace{-0.15in}
    \label{fig:Random_drop_results}
\end{figure}

%\vspace{-0.1in}
\vspace{0.05in}\noindent\textbf{Random Drop Methods}
The random drop strategy is to avoid the optimization being dominated by image features. As shown in Fig.~\ref{fig:No_random_drop}, without using random drop in the training process, the results are poor in areas not covered by image features (the right side of the sofa). This region can only rely on neural 3D representation for rendering; thus, the poor results imply that the neural 3D representations are not well optimized. 
In contrast, our method with the random drop strategy produces high-quality images. 
Moreover, we observe that the results are slightly different when using different variants of the random drop strategy. For example, the rendered image is automatically enhanced using query-point-based random drop, as displayed in  Fig.~\ref{fig:Random_drop_results}. One possible explanation is that, during training, the use of volume rendering for aggregation automatically enhances query points with image features to compensate for query points with low-quality neural 3D features on the same ray.
% when the volume rendering module aggregates two types of features on a ray, the stronger one (\ie, with image features) is automatically enhanced to compensate for the weaker one (\ie, without image features). 
However, this enhancement tends to change the color tone, as demonstrated in the bicycle example in Fig.~\ref{fig:Random_drop_results}. 
% in Fig.~\ref{fig:Random_drop_results}. 
Thus, we currently adopt the ray-based random drop to render images having a closer appearance to the reference images.   

\section{Conclusion}
In this paper, we present an approach to render high-fidelity and view-consistent images in large-scale scenes from sources contaminated by motion blurs.  
%Our method improves the rendering quality from two aspects, including the design of the neural rendering model and the quality of captured data. Specifically, we have
We develop a hybrid neural rendering model that makes use advantages of both image-based representation and neural 3D representation to render high-quality and view-consistent results. We also propose to efficiently simulate blur effects on the rendered image and design a quality-aware training strategy to down-weight blurry images, which helps the hybrid neural rendering model learn from blur-free supervisions and generate high-fidelity images. We conduct experiments on both real and synthetic data and obtain superior performance over previous methods.   

\vspace{0.05in}\noindent\textbf{Limitations} {Our method focuses on dealing with simple motion blurs in the training data, and defocus blur is not considered. Moreover, there are many other in-the-wild challenges, such as images captured under different exposure times and light conditions that require further research.} 
% \xjqi{please move this to supplement if pages are not allowed}

\vspace{0.05in}\noindent\textbf{Acknowledgement} This work has been supported by Hong Kong Research Grant Council - Early Career Scheme (Grant No. 27209621), General Research Fund Scheme (Grant no. 17202422), and RGC matching fund scheme. 
% (RMGS). %Part of the described research work is conducted in the JC STEM Lab of Robotics for Soft Materials funded by The Hong Kong Jockey Club Charities Trust.

%%%%%%%%% REFERENCES
{\small
\bibliographystyle{ieee_fullname}
% \bibliography{egbib}

}

\clearpage
% \title{Supplementary Material for: \\
% ``Hybrid Neural Rendering for Large-Scale Scenes with Motion Blur" } 
% % Using Hybrid Neural Radiance Fields and Blur Simulations}
% \maketitle
\centerline{\Large \textbf{Appendix}}
% In this file, we first introduce more implementation details on network structures, the process of generating synthetic data, and the content-aware blur detection in Sec.~\ref{sec: More_details}. Then, we explore model capacity in Sec.~\ref{sec:model_capacity}. At last, we conduct more experiments and present more results on challenging scenarios and more datasets in Sec.~\ref{sec:more_resuls}.
\vspace{0.2in}
\noindent\textbf{Overview} This document begins by introducing an extendable scheme for predicting the degradation kernel in Section~\ref{sec: Extendable}. We then present additional implementation details on network structures, the generation process of synthetic data, and content-aware blur detection in Section~\ref{sec: More_details}. At last, we investigate model capacity in Section~\ref{sec:model_capacity} before conducting additional experiments and presenting more results on challenging scenarios and diverse datasets in Section~\ref{sec:more_resuls}.

\section{Learnable Degradation Kernel}
\label{sec: Extendable}
\begin{figure}[h]
    \centering
    \includegraphics[width=1.0\linewidth]{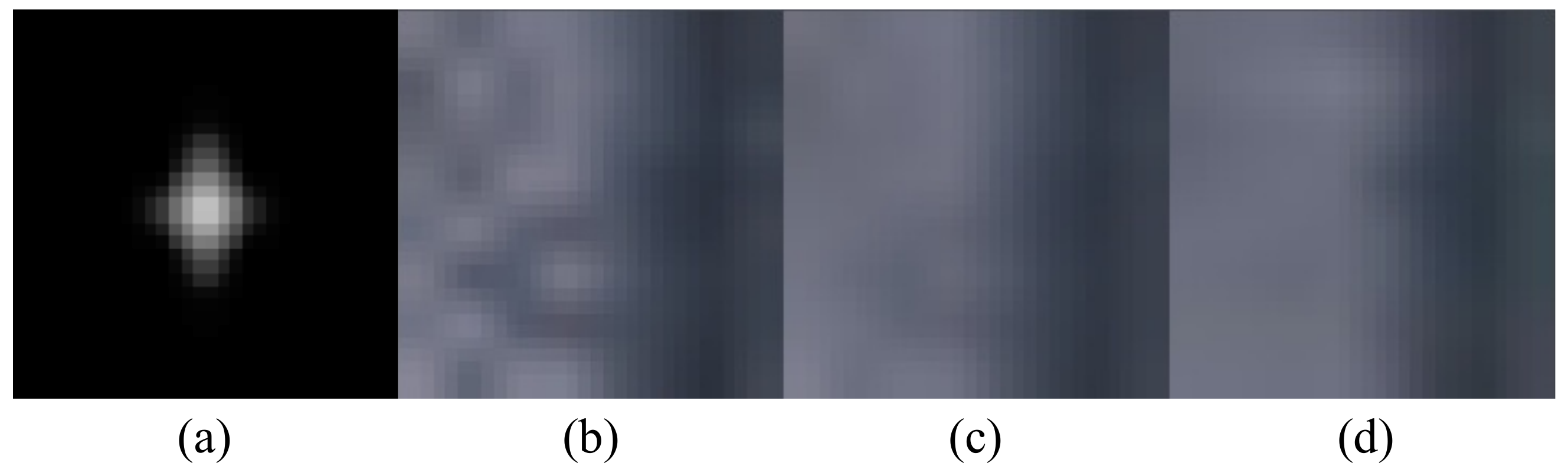}
    \caption{Visualization of learnable degradation kernel. (a) The predicted degradation kernel. (b) The rendered image patch. (c) Apply the degradation kernel to the rendered image patch. (d) The blurry reference image patch.} 
    \label{fig:learned_kernel}
    \vspace{-0.1in}
\end{figure}

\noindent\textbf{Method}
By comparing the appearance of rendered and reference patches, it is possible to predict the potential degradation kernel $B_\text{final}$ that makes up for their difference. To comprehensively explore the patches and efficiently predict the degradation kernel, we adopt a shallow MLP here. More specifically, the rendered and reference patches are first grayscaled and flattened, and then they are concatenated and fed into an MLP to predict a kernel $B_p$ and a blending weight $\omega_p$. At last, the degradation kernel $B_\text{final}$ is obtained by blending the predicted kernel $B_p$ and the identity kernel $B_I$ consisting of zeros except for the center point, which is set as one:
\begin{equation}
    {B}_{final} = \omega_{p}*B_{p} + (1-\omega_{p})*B_{I}.
    \label{eq:leanable_kernel}
\end{equation}
After this, we can directly apply the final degradation kernel to the rendered patch to avoid the negative influence of defective training data (see Fig.~\ref{fig:learned_kernel}).

\vspace{0.1in}
\noindent\textbf{More discussions}
1) To handle view-dependent inconsistencies, previous methods need to optimize a latent code for each training view, and the training time is proportional to the number of training views since each view must be exhaustively sampled to well optimize the corresponding latent code. In contrast, our method directly studies generalizable cues in image patches or uses predefined kernels, and thus is less limited by the number of training views, especially in large-scale scenes covered by dense views. In the future, how to integrate image cues and the per-view latent code optimization is worth exploring. 2) Even though our method focuses on per-scene optimization, the patch/image-based design would enjoy the benefits of pre-training~\cite{wang2021ibrnet,yu2021pixelnerf}. 3) The learnable degradation kernel is very flexible and thus may not be limited to blur artifacts. For example, we can try to use the rendered and reference patches to predict the alignment kernel, where each position represents a motion vector to account for misalignments caused by rough geometries, incorrect camera calibrations, and image distortions. 

\vspace{0.1in}
\noindent\textbf{Results}
We validate the effectiveness of learned degradation kernels by conducting experiments on the ``LivingRoom" dataset using the point-based framework. When compared to the results of pre-defined degradation kernels, the learned degradation kernel achieves 0.5 improvements on PSNR, as shown in Table~\ref{tab:learned_kernels}. As for efficiency, it takes about 10 hours to train (200K iterations) when using only hybrid neural rendering, and an additional 0.5 hours to learn the degradation kernels.

\begin{table}[h]
    \centering
    \resizebox{0.8\columnwidth}{!}{
    \begin{tabular}{c|ccc}
    \hline
    LivingRoom & PSNR$\uparrow$ & SSIM$\uparrow$ & LPIPS$\downarrow$ \\
    \hline
    Pre-defined kernels & 35.30 & 0.980 & 0.051\\
    Learned kernels & \textbf{35.81} & \textbf{0.981} & \textbf{0.050}\\
    \hline
    \end{tabular}}
    \caption{Results of using different degradation kernels.}
    \vspace{-0.1in}
    \label{tab:learned_kernels}
\end{table}

In Fig.~\ref{fig:learned_kernel} (a), we visualize the learned degradation kernel. When this kernel is applied to the rendered sharp image patch (Fig.~\ref{fig:learned_kernel} (b)), the resulting degraded image patch (Fig.~\ref{fig:learned_kernel} (c)) more closely resembles the blurry reference image patch (Fig.~\ref{fig:learned_kernel} (d)), allowing for the preservation of high-frequency information.  

\section{More Implementation Details}
\label{sec: More_details}
\noindent\textbf{Network Architectures}
Our network consists of three major components. \emph{1) The image feature extractor ($G$).} It extracts multi-scale image features using $3\times3$ kernels. We double the intermediate channels when down-sampling with a stride of 2. Totally, it has six layers with intermediate channels of \{6, 6, 12, 12, 24, 24\}. \emph{2) MLPs in the feature aggregation.} In Figure 3 of the main paper, the blue MLP consist of three layers with intermediate channels of \{128, 128, 128\}, and the green MLP has four layers with intermediate channels of \{64, 64, 64, 1\}, and the yellow MLP has three layers with intermediate channels of \{45, 45, 45\}.  \emph{3) MLPs for color and volume density prediction.} In Figure 2 of the main paper, the purple MLP used to predict volume density has five layers with intermediate channels of \{256, 256, 256, 256, 1\}. Additionally, the black MLP that generates color from the hybrid feature has one layer with three output channels.   

\vspace{0.1in}
\noindent\textbf{Synthetic Data}
The $480\times640$ RGB-D image sequences synthesized from Habitat-sim~\cite{habitat19iccv, szot2021habitat, straub2019replica} are sharp. Following the train and test split on ScanNet, we divide the entire dataset into training sets (every 5th image) and testing sets. We then simulate motion blurs~\cite{LeviBorodenko20} on the training sets (RGB images) to obtain blurry training images. Specifically, each image has a probability of 0.75 of being blurred using a motion blur simulator~\cite{LeviBorodenko20}. The simulator has two hyperparameters: size and intensity, which control the properties of the motion blur. The size parameter $k$ controls the moving distance, and is uniformly sampled from the range $k\in(3, 16)$. The intensity parameter $\phi$ determines the level of shake in the moving direction. As we assume that the camera moves in one direction at a given moment while capturing high frame rate videos, we randomly choose a small intensity value of $\phi\in (0, 0.1)$. Note that our model is trained using randomly sampled small patches; therefore, this blur simulation process operates appropriately on small patches.

\vspace{0.1in}
\noindent\textbf{Content-Aware Blur Detection}
To calculate the blurriness score, we use a method called ``variation of the Laplacian"~\cite{pech2000diatom}. First, we apply a blur kernel (\ie, mean filter in our paper) to the original RGB image $I$ to reduce the influence of noise. Then, we use the Laplacian operator to extract the high-frequency components $\hat{H}$ from the denoised image $\hat{I}$. Finally, we obtain the blurriness score by computing the variation of all high-frequency components $S = var(\hat{H})$.

To prevent the inaccurate blurriness score of one frame from affecting the subsequent frames, we implement our content-aware detection (Sec.~3.2 in the main paper) using a sliding window scheme in practice. Specifically, the sliding window incorporates 10 frames (\ie, $N=10$) to calculate quality-aware weights $\omega_i^b$, and moves in steps of 5 on the time axis. If a frame is covered multiple times by the sliding window, the final quality-aware weight is the average of all weights belonging to that frame. In this paper, we use consecutive frames since videos are provided. Similarly, neighboring frames can also be used to compute overlapping regions in place of consecutive frames.

\section{Model Capacity}
\label{sec:model_capacity}
To further confirm that the improvements of our method, which builds upon Point-NeRF, are not dependent on larger model capacities, we conducted two experiments on Point-NeRF: \textit{1) Gradually increasing the number of points from 4.2 million to 7.4 million, and 2) Increasing the channels of each point descriptor from 32 to 63.} Results in Table~\ref{tab:num_params} indicate that naively increasing the capacity of Point-Nerf does not improve performance. Moreover, the total number of parameters optimized in Point-NeRF with 7.4 million points (model size: 1.2 GB) already exceeds that of our hybrid model with 4.2 million points (model size: 682 MB, PSNR: 31.25). 
% And the parameters of other methods are shown in Table.~\ref{tab:num_params}.

\begin{table}[h]
    \centering
    \resizebox{1.0\columnwidth}{!}{
    \begin{tabular}{c|ccccc}
    \hline
    Number of points & 4.2M & 5.3M & 6.3M & 7.4M & 4.2M\\
    Number of channels & 32 & 32 & 32 & 32 & 63\\
    \hline
    %  & \multicolumn{5}{c}{Number of points + Number of channels}\\
    %  & 4.2M+32 & 5.3M+32 & 6.3M+32 & 7.4M+32 & 4.2M+63\\
    % \hline
    Point-NeRF (PSNR) & 30.54 & 30.56 & 30.55 & 30.50 & 30.58\\
    \hline
    \end{tabular}}
    \caption{{Results on ``Scene241\_01" while increasing the number of point descriptors and channels of each descriptor. Naively increasing the capacity of Point-NeRF cannot boost performance.}}
    \vspace{-0.1in}
    \label{tab:num_params}
\end{table}

% \begin{table}[h]
%     \centering
%     \resizebox{0.75\columnwidth}{!}{
%     \begin{tabular}{c|cc}
%     \hline
%     Parameters & Network & Descriptors $\&$ Channels\\
%     \hline
%     IBRNet & 0.04M & --- \\
%     Deblur-NeRF & 1.3M & --- \\
%     NPBG & 1.96M & 5.8M $\&$ 8 \\
%     Point-NeRF & 0.34M & 4.2M $\&$ 32 \\
%     Ours & 0.45M & 4.2M $\&$ 32  \\
%     \hline
%     \end{tabular}}
%     \caption{\peng{Number of trainable parameters in the per-scene optimization process.\peng{Do we need this?}}}
%     \vspace{-0.1in}
%     \label{tab:ab_features}
% \end{table}

\begin{figure}[h]
    \centering
    \includegraphics[width=0.95\linewidth]{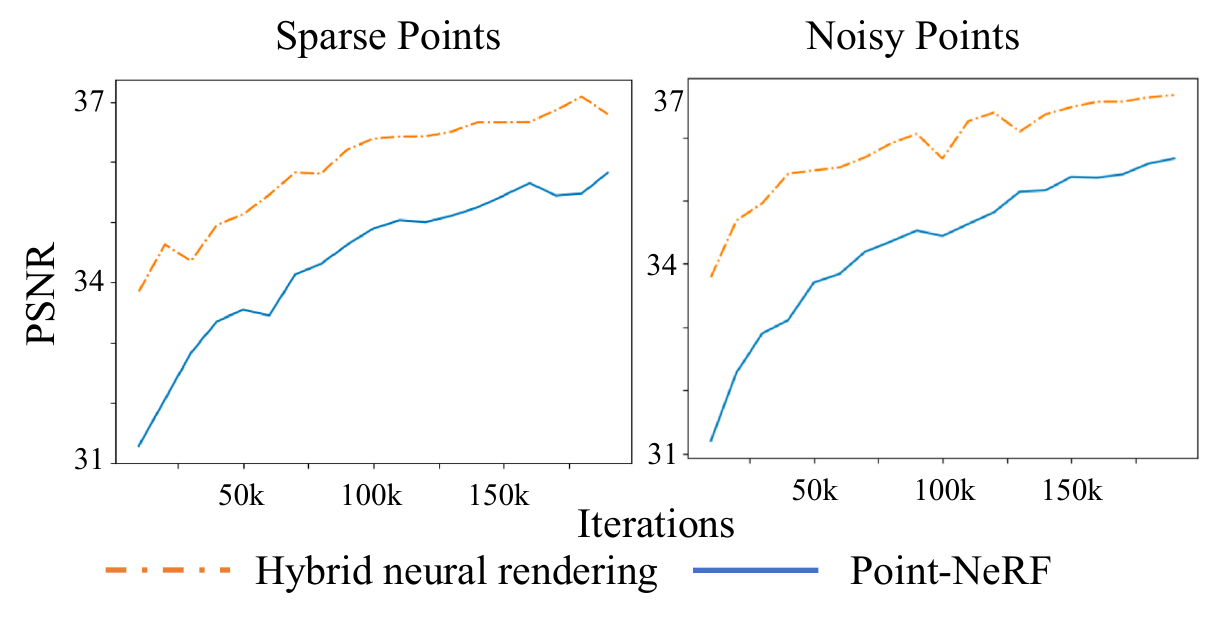}
    \caption{The trend of PSNR across different training iterations is shown for the case of sparse points and noisy points. When using image features (\ie, hybrid neural rendering), the performance is better than the neural-3D-feature-only design (\ie, Point-NeRF).} 
    \label{fig:Ab_imageFeats}
    \vspace{-0.1in}
\end{figure}

\section{More Experiments}
\label{sec:more_resuls}
\noindent\textbf{Sparse and Noisy Points}
We conduct experiments on challenging situations using sparse (0.6M points) and noisy (add gaussian noise $\mathcal{N}(0, 0.05)$) points on 'VangoRoom' (to validate the effectiveness of image features, all blur-related designs are disabled, and images used for training and testing are blur-free). As shown in Fig.~\ref{fig:Ab_imageFeats}, using image features converges faster and achieves better performance than Point-NeRF, which only utilizes neural 3D features. 

\begin{figure}[h]
    \centering
    \includegraphics[width=1.0\linewidth]{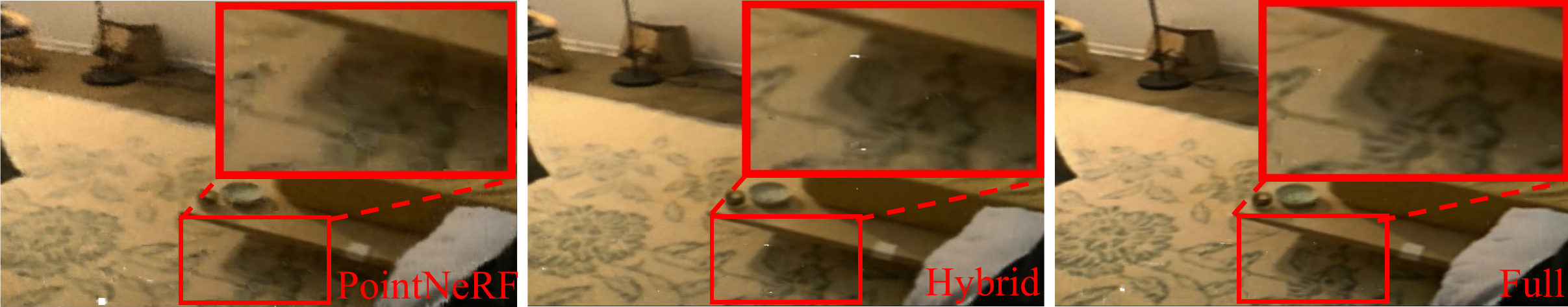}
    \caption{Qualitative results on the ARKITScenes dataset. Each component of our proposed method contributes to the sharpness of rendered images.}
    \label{fig:arkits_dataset}
    \vspace{-0.1in}
\end{figure}

\vspace{0.1in}
\noindent\textbf{ARKITScenes Dataset}
{We present results on the ARKITScenes dataset (``Scene\_40776204")~\cite{dehghan2021arkitscenes}. Fig.~\ref{fig:arkits_dataset} demonstrates the effectiveness of our hybrid rendering and blur-handling designs, where all components contributing to the sharpness of rendered images (note the bird). Quantitative results in Table~\ref{tab:arkitscenes} indicate that our method outperforms Point-NeRF on all three metrics.}

\begin{table}[h]
    \centering
    \resizebox{0.8\columnwidth}{!}{
    \begin{tabular}{c|ccc}
    \hline
    ``Scene\_40776204" & PSNR$\uparrow$ & SSIM$\uparrow$ & LPIPS$\downarrow$\\
    \hline
     Point-NeRF & 31.02 & 0.947 & 0.212\\
     Ours(H) & \textbf{32.55} & \textbf{0.961} & \textbf{0.127}\\
     Ours(Full) & 31.97 & 0.957 & 0.135\\
    \hline
    \end{tabular}}
    \caption{Quantitative results on the ARKITScenes dataset.}
    \vspace{-0.1in}
    \label{tab:arkitscenes}
\end{table}

\begin{figure}[htb!]
    \centering
    \includegraphics[width=0.9\linewidth]{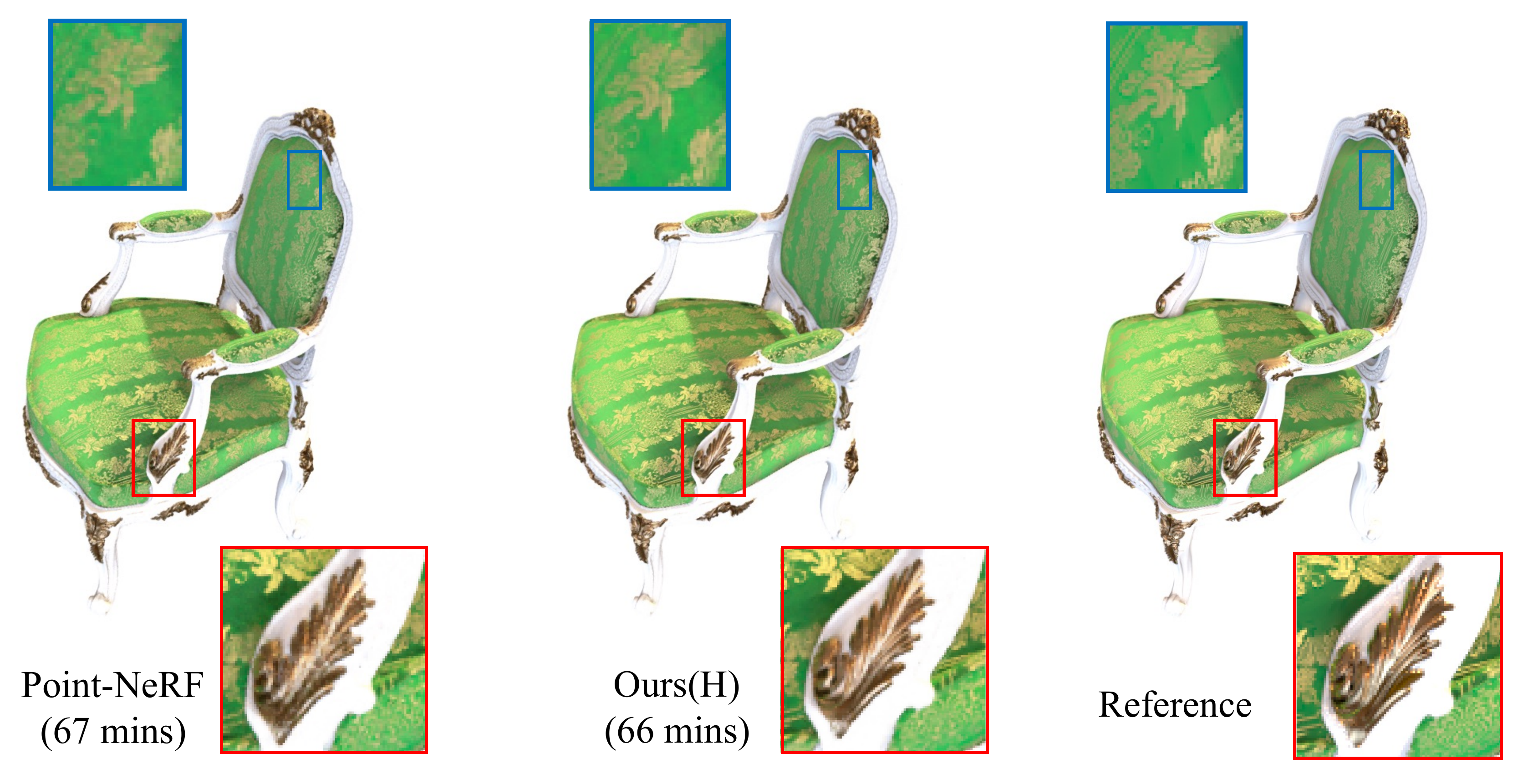}
    \caption{Qualitative results on the NeRF synthetic data. Leveraging image features assists in high-frequency details generation.}
    \label{fig:nerf_dataset}
    \vspace{-0.1in}
\end{figure}

\vspace{0.1in}
\noindent\textbf{NeRF Synthetic Dataset}
{We provide results on the NeRF synthetic dataset (``Chair")~\cite{mildenhall2021nerf}, which only provides posed RGB images without depth information. When trained for 66 minutes, our method leveraging image-based features produces results with more high-frequency details, as displayed in Fig.~\ref{fig:nerf_dataset}. This observation is also corroborated by quantitative results presented in Table~\ref{tab:nerf_syn}.} 

\begin{figure*}[htb!]
    \centering
    \includegraphics[width=1.0\linewidth]{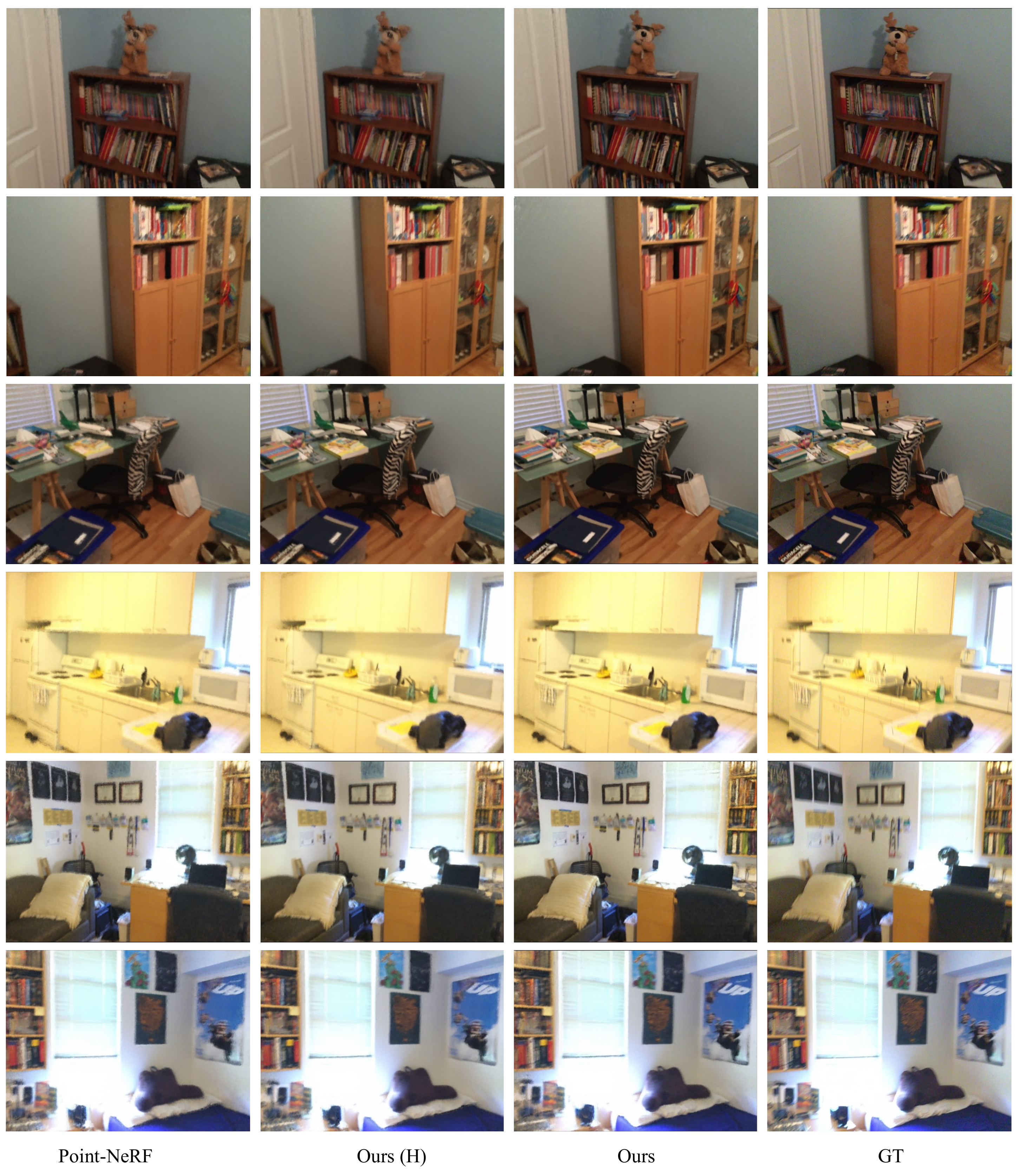}
    \caption{More results compared to Point-NeRF.}
    % Point-NeRF: it suffers from noisy edges and blurry outputs. Ours (H): it contains more details and improves the smoothness. Ours: the results are sharper while dealing with blurriness. (Please zoom in for details.) 
    \label{fig:more_results}
\end{figure*}

\vspace{0.1in}
\noindent\textbf{ScanNet Dataset}
In Fig.~\ref{fig:more_results}, we show more qualitative results and compare them with Point-NeRF to demonstrate the superiority of our approach. We observed that the outputs of Point-NeRF suffer from blurry outputs and noisy edges, while ``Ours (H)" contains more details and smoother appearance, evident in the toy plane in the third row and characters on the posters in the last two rows. Moreover, the image sharpness is further improved while applying our designs to handle blur artifacts.

\vspace{0.1in}
\noindent\textbf{Video Results}
In the video results, we compare our approach with other baselines, including NeRF~\cite{mildenhall2021nerf}, IBRNet~\cite{wang2021ibrnet}, NPBG~\cite{aliev2020neural}, Point-NeRF~\cite{xu2022point} and Deblur-NeRF~\cite{ma2022deblur}, to demonstrate the superiority of our approach in terms of quality and consistency. Moreover, we also validate the effectiveness of our designs, including hybrid rendering, handling blurriness, random drop, and neural 3D features.

\begin{table}[h]
    \centering
    \resizebox{0.9\columnwidth}{!}{
    \begin{tabular}{c|ccc}
    \hline
    ``Chair" & PSNR$\uparrow$ & SSIM$\uparrow$ & LPIPS$\downarrow$\\
    \hline
     $\text{Point-NeRF}_{40k}$ (67 mins) & 33.80 & 0.986 & 0.017\\
     $\text{Ours (H)}_{20k}$ (66 mins) & 34.40 & 0.988 & 0.016\\
    \hline
     $\text{Point-NeRF}_{200k}$ & 35.70 & 0.992 & 0.010\\
     $\text{Ours (H)}_{200k}$ & \textbf{36.23} & \textbf{0.993} & \textbf{0.009}\\
    \hline
    \end{tabular}}
    \caption{Quantitative results on the NeRF synthetic dataset. Our method outperforms Point-NeRF in both scenarios where training is performed for the same amount of time and the same number of iterations.}
    \vspace{-0.1in}
    \label{tab:nerf_syn}
\end{table}

\end{document}